\documentclass[a4 paper,12pt]{article}
\usepackage{graphicx}
\usepackage{amsmath}
\usepackage{authblk}
\usepackage[utf8]{inputenc}
\usepackage[english]{babel}
\usepackage{csquotes}
\usepackage{amssymb}
\usepackage{caption}
\usepackage[margin=2.5cm]{geometry}
\usepackage{setspace}
\usepackage{float}

\usepackage{algpseudocode}
\usepackage{algorithm}
\usepackage{algorithmicx} 
\usepackage{amsthm}
\usepackage{mathtools}
\usepackage[toc,page]{appendix}
\usepackage{tabulary}
\usepackage{booktabs}
\usepackage{subcaption}
\usepackage[authoryear, round]{natbib}
\usepackage{mathtools}

\usepackage[colorlinks = true, pdfstartview = FitV, linkcolor = blue, citecolor = blue, urlcolor = blue, bookmarks = false]{hyperref}
\usepackage{pdflscape}
\usepackage{afterpage}
\usepackage{capt-of}
\usepackage{bbm}
\usepackage{colortbl}
\usepackage{url}
\usepackage{bbold}
\usepackage{chngcntr}
\counterwithin{equation}{section}

\usepackage{authblk}
\usepackage[export]{adjustbox}
\usepackage{cleveref}
\usepackage[most]{tcolorbox}
\newtcbtheorem{Remark}{Remark}{
  enhanced,
  sharp corners,
  attach boxed title to top left={
    yshifttext=-1mm
  },
  colback=white,
  colframe=blue!75!black,
  fonttitle=\bfseries,
  boxed title style={
    sharp corners,
    size=small,
    colback=blue!75!black,
    colframe=blue!75!black,
  } 
}{remark}

\theoremstyle{definition}

\theoremstyle{remark}

\title{Preconditioned Neural Posterior Estimation for Likelihood-free Inference}
\author{Xiaoyu Wang\textsuperscript{1,2*}, Ryan P. Kelly\textsuperscript{1,2}, David J. Warne\textsuperscript{1,2}, Christopher Drovandi\textsuperscript{1,2}
\\
\textbf{1} School of Mathematical Sciences, Queensland University of Technology, Brisbane, QLD, Australia
\\
\textbf{2} Centre for Data Science, Queensland University of Technology, Brisbane, QLD, Australia\\
\textbf{*} Email: x311.wang@hdr.qut.edu.au}

\begin{document}
\maketitle

\begin{abstract}
    Simulation based inference (SBI) methods enable the estimation of posterior distributions when the likelihood function is intractable, but where model simulation is feasible.  Popular neural approaches to SBI are the neural posterior estimator (NPE) and its sequential version (SNPE).  These methods can outperform statistical SBI approaches such as approximate Bayesian computation (ABC), particularly for relatively small numbers of model simulations.  However, we show in this paper that the NPE methods are not guaranteed to be highly accurate, even on problems with low dimension.  In such settings the posterior cannot be accurately trained over the prior predictive space, and even the sequential extension remains sub-optimal. To overcome this, we propose preconditioned NPE (PNPE) and its sequential version (PSNPE), which uses a short run of ABC to effectively eliminate regions of parameter space that produce large discrepancy between simulations and data and allow the posterior emulator to be more accurately trained.  We present comprehensive empirical evidence that this melding of neural and statistical SBI methods improves performance over a range of examples, including a motivating example involving a complex agent-based model applied to real tumour growth data.

\end{abstract}

\section{Introduction}

Computational models, frequently termed as simulators, are typically governed by stochastic processes. When provided with a set of parameter values, these simulators output synthetic data that inherently capture the stochastic nature of the simulated phenomena.  However, a substantial challenge arises in performing posterior inference for the parameters of these simulators, as the corresponding likelihood function is often intractable.  An example is agent-based modelling of tumour growth (e.g.\ \citet{jenner2020enhancing,aylett2021june}), where cell proliferation, movement and invasion are governed by probabilistic rules, which depend on key biological parameters. Consequently,  standard statistical inference methods that rely on a closed form expression for the likelihood function are inapplicable in this scenario. 

To address this issue, simulation based inference (SBI) methods have been developed, which approximate the posterior based only on simulations from the model.  The most thoroughly examined SBI method in the statistical literature is approximate Bayesian computation (ABC) \citep{sisson2018handbook}.  Advancements in deep neural networks have led to the emergence of neural SBI methods \citep{cranmer2020frontier}. Their widespread application spans various fields, including biology \citep{wehenkel2023simulation}, neuroscience \citep{fengler2021likelihood,west2021inference},  and astronomy \citep{mishra2022inferring,dax2021real}. 

Statistical SBI methods, such as ABC methods, are well-developed and boasts strong theoretical guarantees of convergence to the true posterior \citep{beaumont2009adaptive,blum2010approximate,biau2015new,lintusaari2017fundamentals,sisson2018handbook,beaumont2019approximate}. ABC approaches compare observed to simulated data using a discrepancy function and prefer parameter values that generate discrepancies below a pre-defined threshold, $\epsilon$.  However, often a small value of $\epsilon$ is required to obtain accurate posterior approximations, which can significantly increase the number of model simulations required, and hence the computational cost \citep{csillery2010approximate,turner2012tutorial}.  Advanced ABC samplers such as adaptive sequential Monte Carlo ABC (SMC ABC, e.g.\ \citet{sisson2007sequential,drovandi2011estimation}) have been developed to mitigate this issue. However, the efficiency of ABC methods may decrease as $\epsilon$ becomes smaller, requiring a significantly larger number of simulated datasets. Enhancing the efficiency of ABC algorithms remains an active area of research \citep{lintusaari2017fundamentals,sisson2018handbook}. 

With the rapid advances in machine learning methods, more efficient approaches based on neural networks have been developed \citep{papamakarios2017masked,papamakarios2021normalizing}. A popular approach is neural posterior estimator (NPE) and its sequential version, the sequential NPE (SNPE) \citep{papamakarios2016fast, lueckmann2017flexible,greenberg2019automatic}.  These methods use a set of training pairs of parameter values and simulated datasets to fit a neural conditional density estimator (NCDE), such as a conditional normalising flow \citep{rezende2015variational,winkler2019learning,durkan2019neural,dolatabadi2020invertible}, to approximate the posterior.  NPE uses parameter values drawn from the prior, whilst SNPE uses parameter values drawn from previous NPE approximations for a given number of rounds.  The idea of the sequential approach is that a more accurate emulator of the posterior can be achieved when more (parameter value, simulated data) training pairs are generated in higher density regions of the posterior. \citet{lueckmann2021benchmarking} have shown that NPE can outperform ABC, particularly for a relatively small number of model simulations.

However, in real-world problems, there may be little known about the parameters \emph{a priori}, so that a vague prior may be employed. For example, a uniform distribution with a wide constraint range may be used as a prior distribution. For some problems in this setting we find that it is difficult to construct an accurate NCDE across a wide parameter space, which leads to NPE producing an inaccurate posterior approximation.  We find that even SNPE may not able to recover from such an initially deficient approximation, even with a relatively large number of rounds, and hence model simulations.  One way to mitigate this kind of unstable NCDE training is to clip the extreme simulated datasets \citep{shih2023fast, de2023field}.  However, this approach is ad-hoc and it is not clear how much clipping is required for a given problem and may require extensive experimentation, and each level of clipping requires refitting of the NCDE.  

This paper contains three key contributions.  Firstly, we explore several examples where NPE methods fail to produce highly accurate posterior approximations, even in relatively low dimensional problems.  Our second contribution is the development of preconditioned NPE (PNPE), and its sequential extension (PSNPE), which combines the strengths of statistical and ML approaches to SBI.  The preconditioning step involves applying an ABC algorithm for efficiently discarding parts of the parameter space that lead to large discrepancies, which then subsequently permits NPE methods to perform well.  In a sense, our preconditioning step acts as a principled clipping method.  Our third contribution shows via an extensive empirical study that our preconditioned NPE approaches outperform NPE approaches when the latter performs sub-optimally, and is competitive when it performs well. Our motivating example involves fitting a complex agent-based model of tumor growth to real cancer data.  

\section{Simulation-based Inference}

Consider a simulator that takes parameters $\theta \in \mathbb{R}^d$ where $d$ is the number of parameters and generates a simulated dataset $x \in \mathbb{R}^D$ where $D$ is the dimension of the data, but its density $p(x|\theta)$ is intractable. The objective of SBI is to accurately estimate the posterior density of $\theta$ conditional on the observed dataset $x_o \in \mathbb{R}^D$ based only on simulating data from the model and not requiring evaluation of the intractable likelihood, $p(x_o|\theta)$.  Two popular SBI methods are ABC and NPE, which are summarised below.

\subsection{Approximate Bayesian computation}

Statistical SBI \citep{sisson2018handbook}, such as the ABC rejection algorithm, is based on Monte Carlo rejection sampling.  That is, it keeps only the parameter values simulated from the prior that generate simulated data $x$ such that $\rho(x,x_o) < \epsilon$, where $\rho(x,x_o)$ is a user-defined discrepancy function between the simulated and observed data, and $\epsilon$ is a user-defined threshold often referred to as the ABC tolerance.  

SMC ABC algorithms (e.g.\ \citet{sisson2007sequential,drovandi2011estimation}) aim to be more efficient by sampling a sequence of ABC posteriors with decreasing $\epsilon$'s, updating the importance distribution at each iteration. More specifically, SMC ABC algorithms define a sequence of non-increasing ABC thresholds $\epsilon_1 \geq \epsilon_2 \geq \dots \geq \epsilon_T$, such that
\begin{equation}
p_{\epsilon_t}(\theta|x_o) \propto p({\theta})\int_{\mathbb{R}^D} \mathbb{I}\big( \rho(x_o,x) < \epsilon_t \big)p(x{\mid}{\theta}) dx, \qquad \text{for } t = 1,\dots,T.
\end{equation}
Here, $\epsilon_T = \epsilon$ represents the target ABC posterior. 

In many real-world applications, $x, x_o \in \mathbb{R}^D$ are considered high-dimensional data, necessitating a mapping to a lower-dimensional space for computational efficiency. This is typically done using summary statistics $S(\cdot)$. If summary statistics are required, we use $S(x)$ and $S(x_o)$ instead of the full datasets $x$ and $x_o$. The choice of appropriate summary statistics is a subject of ongoing research and is discussed extensively in the literature (see \citet{sisson2018handbook}). 

However, even sophisticated ABC algorithms can require a significant number of model simulations to achieve a suitably small value of $\epsilon$ \citep{biau2015new,csillery2010approximate,beaumont2009adaptive,blum2010approximate}.

\subsection{Neural posterior estimation}
NPE uses $N$ training pairs of simulator parameter values and simulated datasets, ${\{\theta_i, x_i\}}_{i=1}^N$, to estimate the posterior distribution $p(\theta|x)$ \citep{papamakarios2021normalizing}. Once the NPE is trained on the simulated datasets, the posterior distribution $p(\theta|x_o)$ can be computed by inputting the observed dataset $x_o$. A conditional neural density estimator $q_{F(x,\psi)}(\theta)$, utilizing a neural network $F$ and its adjustable network weights $\psi$, is often used as an NPE. In order to train $q_{F(x,\psi)}(\theta)$, the following loss is minimized:

A neural density estimator $q_{F(x,\phi)}(\theta)$ used neural network $F$ with adjustable network weights $\phi$ is used for NPE that trained by minimize the loss:
\begin{equation}
    \psi^* = \mathop{\arg \min}\limits_{\psi} -\sum_{i = 1}^N \mathrm{log} q_{F(x_i,\psi)}(\theta_i),
\end{equation}
over network weights $\psi$. For a sufficiently  expressive $q_F$, $q_{F(x,\psi)}(\theta) = p(\theta|x)$. 

SNPE aims to improve the accuracy of the approximate posterior for a particular observed dataset $x_o$ iteratively by sampling parameter values from a previous NPE approximation for a given number of rounds. The current NPE approximation is treated as a proposal distribution $\Tilde{p}(\theta)$ for the next round. However, training $q_F$ using parameter values drawn from $\Tilde{p}(\theta)$ will not converge to the true posterior distribution, but rather to
\begin{equation}
\label{npe equation}
\Tilde{p}(\theta|x_o) \propto p(\theta|x_o)\frac{\Tilde{p}(\theta)}{p(\theta)}.
\end{equation}

Many approaches have been developed to overcome this limitation, such as \citet{papamakarios2016fast,lueckmann2017flexible,greenberg2019automatic}. Among all these approaches, we use the automatic posterior transformation (APT, also known as SNPE-C), as proposed by \citet{greenberg2019automatic}, which has been reported to significantly outperform the others \citep{lueckmann2021benchmarking}. However, APT  can suffer from a `leakage' issue, in which case we use truncated sequential neural posterior estimation (TSNPE) \citep{deistler2022truncated}, which aims to overcome the leakage problem.   For simplicity, we  refer to our specific implementation as SNPE hereafter, but we note that other implementations of SNPE can be used with our method.

\subsubsection{Illustrative Example}\label{illustration experiments}
We consider a sparse vector autoregressive (SVAR) model that has been considered previously in the SBI literature  \citep{thomas2020split, drovandi2023improving}. The SVAR model is given by:
\begin{equation}
    y_t = Xy_{t-1} + \xi_t,
\end{equation}
where $y_t \in \mathbb{R}^k$ represents the $k$-dimensional observation of the time series at time $t$, $X \in \mathbb{R}^{k\times k}$ is the transition matrix, and $\xi_t \sim \mathcal{N}(0,\sigma^2\mathbb{I})$ is a $k$-dimensional noise vector with $\sigma^2$ being the noise parameter. The model considers a sparse transition matrix $X$ where the only off-diagonal entries that are non-zero must satisfy the following conditions: (1) they are strictly off-diagonal entries, meaning $i \neq j$ for $1 \leq i,j \leq k$; (2) if variable $i$ is coupled with variable $j$, then $X_{i,j} \neq 0$ and $X_{j,i} \neq 0$ (note that $X_{i,j}$ is not necessarily equal to $X_{j,i}$). To ensure the stability of the SVAR, the diagonal elements of $X$ are set to -0.1. The parameter space of SVAR can easily scale to higher dimensions by increasing $k$. In this study, the model parameters $\theta \in \mathbb{R}^{k+1}$ are the non-zero off-diagonal entries of $X$ and its variance and we consider $k = 6$, which leads to 7 parameters. This choice is based on the assumption that if SNPE does not produce highly accurate approximations in this low-dimensional case, it is unlikely to be accurate in a higher dimensional parameter space. We generate an observed dataset of length $T = 1000$ using the true parameter value $\theta = (0.579, -0.143, 0.836, 0.745, -0.660, -0.254, 0.1)$. We use summary statistics to reduce the dimension of the data.  Following \citet{thomas2020split, drovandi2023improving}, we use the lag 1 autocovariance $\frac{1}{T}\sum_{t=2}^T y_t^iy_{t-1}^j$ as the summary statistics, where $y_t^i$ is the $t$th observation of the $i$th time series. We use the sample standard deviation of the $k$ time series to inform $\sigma$.  Thus there is a single summary statistic that is intended to be informative about each parameter.

We employ a uniform distribution as the prior, constrained between -1 and 1 for the $k$ parameters and between 0 and 1 for $\sigma$.  We find that extreme values of the summary statistics can be produced by parameter values  away from the true parameter value. To stabilize the training, we clip simulated datasets with summary statistic outliers (any simulated values greater than ${10}$, around $3\%$ of training datasets).  Note that some experimentation was required to obtain a clipping value that led to reasonable results for SNPE.

\begin{figure}[H]
\begin{center}
\centerline{\includegraphics[width=1\columnwidth]{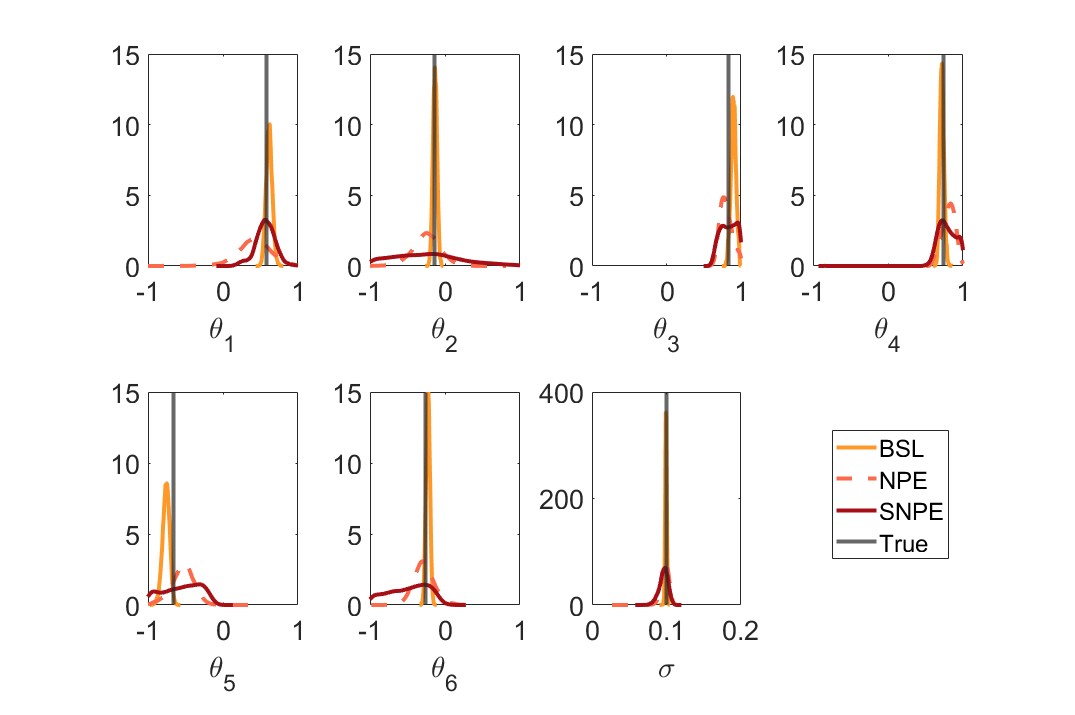}}
\caption{Comparison of marginal posterior distributions between BSL (orange), NPE (dashed pink) and SNPE (red), with black dashed lines representing the true values. The SNPE results are based on three rounds.}
\label{illustration_example}
\end{center}
\end{figure}

\begin{figure}[H]
\begin{center}
\centerline{\includegraphics[width=1\columnwidth]{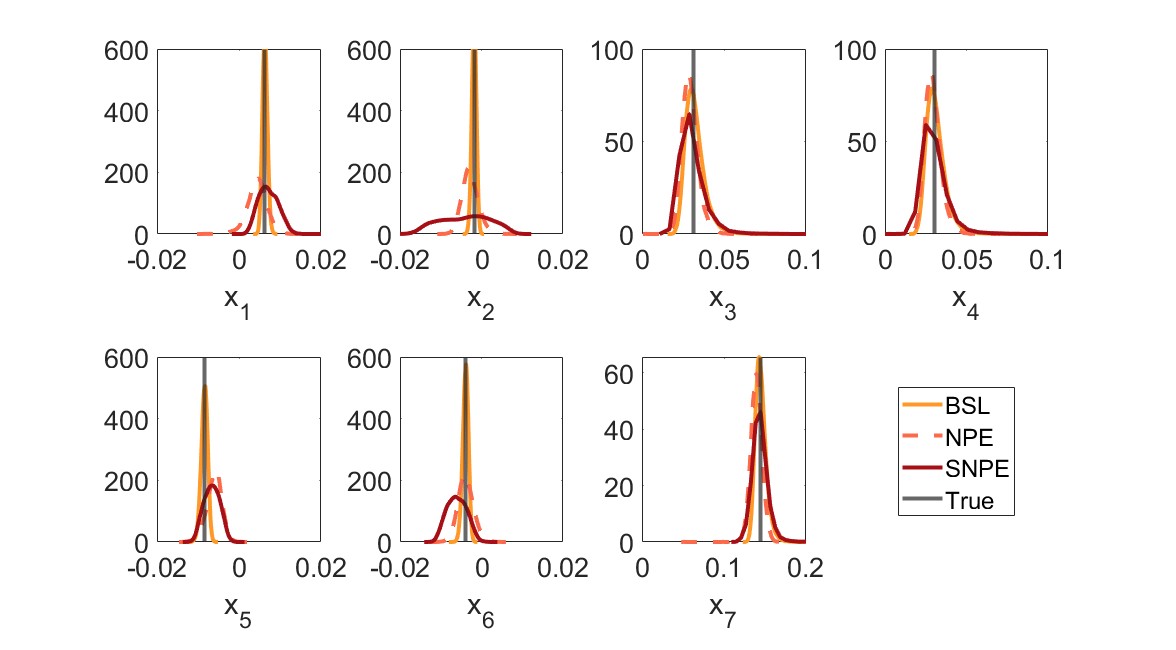}}
\caption{Comparison of posterior predictive distribution of the summary statistics of observation datasets between BSL (orange), NPE (dashed pink) and PNPE (red), with black dashed lines representing the true values. The SNPE results are based on three rounds.}
\label{illustration_example posterior predictive}
\end{center}
\end{figure}

For illustrative purposes, we run three rounds of SNPE and compare the results with those from BSL, considering BSL results as the gold standard for this example. Ideally, we would expect performance to improve when increasing the number of SNPE rounds. However, as shown in Figures \ref{illustration_example} and \ref{illustration_example posterior predictive}, even with datasets clipped for every round, SNPE does not improve the accuracy of the estimates as the number of rounds increases.

\section{Method}

We find that NPE can perform sub-optimally when the prior predictive distribution of the data is complex and has significant variability.  When this occurs, the NCDE may not be sufficiently accurate, especially in regions of high posterior support.  The sequential version of NPE was originally designed to overcome this issue, however if the initial NPE approximation is not substantially better than the prior, then subsequent rounds of SNPE may suffer from the same issue.

Other approaches to overcome this issue may be to increase the training sample size or to try different configurations of the neural network, but both of these may increase the computational cost substantially and may not address the issue.  Instead, we propose the preconditioned NPE (PNPE) method, and its sequential extension below, in order to make NPE methods more reliable.

\subsection{Preconditioned NPE}
For a vague prior distribution $p(\theta)$, parameter values drawn from it might be very far from the true posterior. We suggest using a short run of ABC to refine those parameter values so that they are closer to the true posterior $p(\theta|x_o)$, i.e.,
\begin{equation}
p_{\epsilon}(\theta|x_o) \propto p(\theta)\int_{\mathbb{R}^D} p(x|\theta)\mathbb{I}(\rho(x_o,x) < \epsilon)dx,
\end{equation}
where $\mathbb{I}(\cdot)$ is the indicator function and $\epsilon$ can be chosen considerably larger than what might typically be used in an ABC algorithm. Then we fit a density estimator to those parameter samples. The key idea is to use an efficient ABC algorithm to quickly discard poor regions of the parameter space that generate unusual datasets relative to the observed data, which provides better quality training datasets for NPE. We note that any ABC algorithm could be employed here, but we use the SMC ABC algorithm of \citet{drovandi2011estimation} in this paper (see Appendix \ref{smc-abc details} for a full description of this method).  The SMC ABC algorithm generates $n$ samples from a sequence of ABC posteriors based on decreasing ABC thresholds, $\epsilon_1 > \dots > \epsilon_T$, where $\epsilon_T = \epsilon$ is the target ABC threshold. The sequence of tolerances is determined adaptively, by, at each iteration of SMC, discarding a proportion of the samples, $a\cdot n$, with the highest discrepancy, where $a$ is a tuning parameter.   Then, the population of samples is rejuvenated through a resampling and move step. During the move step, a Markov chain Monte Carlo (MCMC) ABC kernel is employed to maintain the distribution of particles based on the current value of the tolerance. The number of MCMC steps $R_t$ to apply to each particle is determined adaptively based on the overall MCMC acceptance rate, that is $R_t = \left\lceil \frac{\log(c)}{\log(1-p_t^{\mathrm{acc}})}\right\rceil$, where $p_t^{\mathrm{acc}}$ is the estimated MCMC acceptance probability at the $t$th SMC iteration.  A natural stopping rule for the algorithm is when the MCMC acceptance rate becomes intolerably small. 

Based on $n$ parameter samples from the ABC posterior, we fit an unconditional normalising flow $q_{G}$ (note that other density estimators could be used). Then we can use $q_{G}$ as the initial importance distribution for the (S)NPE process.  We call this melding of ABC and (S)NPE as the preconditioned (S)NPE method.  The method is summarised in Algorithm \ref{PNPE}.

\begin{algorithm}[ht]
   \caption{Precondition SNPE}
   \label{PNPE}
\begin{algorithmic}[1] 
    \State Choose preconditioning ABC algorithm and SNPE implementation
    \State Obtain ${\{\theta_i^*\}}_{i=1}^{n}$ from the preconditioning ABC algorithm
    \State Set $\phi^* \leftarrow \mathop{\arg \min}\limits_{\phi} \mathop{\sum}\limits_{i = 1}^{n} -\mathrm{log}q_{G(\phi)}(\theta_i^*)$
    \State Perform SNPE using initial importance distribution $q_{G(\phi^*)}(\theta)$
\end{algorithmic}
\end{algorithm}

If we obtain a well-trained unconditional normalizing flow, this unconditional normalizing flow can act as the initial importance distribution, i.e., $\Tilde{p}(\theta)$  in Equation \ref{npe equation}, and to draw samples for training the NPE.  Following \citet{papamakarios2016fast}, it is noted that given an expressive enough conditional normalizing flow, NPE converges to the true posterior $p(\theta|x_o)$ as $N \rightarrow \infty$, with an appropriate importance re-weight if $\Tilde{p}(\theta) \neq p(\theta)$.

Choosing a suitable value of $\epsilon$ for our method requires some thought.  A smaller value of $\epsilon$ will focus in on more promising regions of the parameter space, but will increase the computational time of the preconditioning step.  A larger value of $\epsilon$ will lead to a fast preconditioning step, but may not eliminate enough of the poor parts of the space to improve the training of the NCDE.  In this paper we use an MCMC acceptance rate of 10\% (unless otherwise specified) as the stopping criteria for the SMC ABC algorithm in the preconditioning step. For our examples we find that this choice is effective at balancing the aforementioned objectives. We note that other choices are possible. 

Furthermore, once these poor simulations have been removed, we find NPE to be more effective than ABC, since ABC requires an exponentially increasing number of simulations to drive $\epsilon$ to 0. To avoid the scaling problem of ABC, the preconditioning step only takes a short run of ABC, and thus we are not interested in driving $\epsilon$ to 0.

\subsection{Computational cost}

We now consider computational cost for P(S)NPE and compare it with SNPE. The preconditioning step can be considered as the initial round of NPE where the total number of simulated datasets generated during SMC ABC is denoted as $n_\mathrm{ABC}$. Hence, it is worth noting that P(S)NPE, like SNPE, is not amortized since it requires running an ABC algorithm for each observation datasets $x_0$.

Furthermore, for complex real-world problems, the simulation time may depend on the parameter values, and  parameter values with very low posterior support can produce substantially longer simulation times.  For such problems, it is important from a computational perspective to quickly eliminate such regions from the parameter space, as is the motivation of the preconditioning ABC algorithm. Thus, for problems where SNPE does not perform well, we find PSNPE to be substantially more computationally efficient in terms of compute time. 


\subsection{Illustrative Example Revisited}
We apply PNPE to the illustrative example shown in Section \ref{illustration experiments}. For the preconditioning step, we use the adaptive SMC ABC algorithm proposed by \cite{drovandi2011estimation}, with tuning parameters $n = 1$k, $a=0.5$, and $c=0.01$. We employ an unconditional normalizing flow as the unconditional density estimator. For this, we use the state-of-the-art neural spline flow implemented in the \texttt{Pyro} package. 

In order to make a fair comparison between P(S)NPE and SNPE, we use the same number of simulations as in the SMC ABC algorithm, denoted $n_{\mathrm{ABC}}$, to train the initial round of NPE. We run the SMC ABC algorithm ten times to obtain the average number of simulations it requires, which is $n_{\mathrm{ABC}} = 54$k. For illustrative purposes, we only run two rounds of SNPE and compare it with PNPE.

\begin{figure}[H]
\begin{center}
\centerline{\includegraphics[width=1\columnwidth]{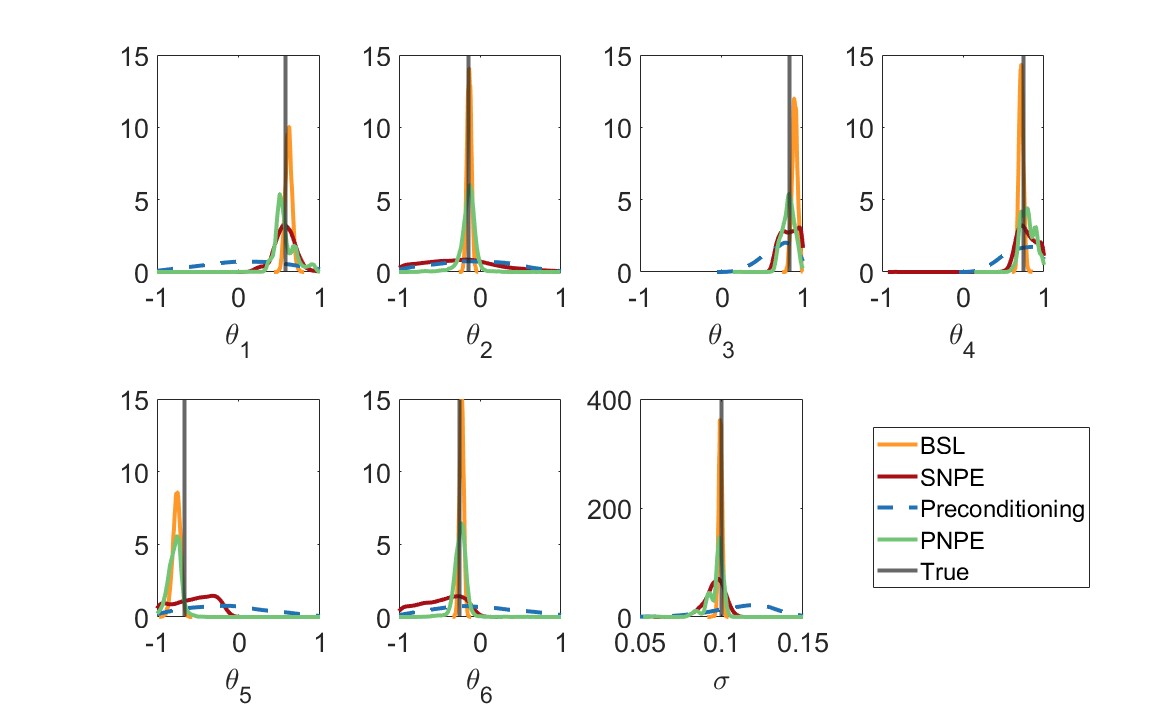}}
\caption{\textbf{Performance on SVAR model.} Comparison of marginal posterior distributions between BSL (orange), SNPE (red), preconditioning step (blue dash) and PNPE (green solid), with black dashed lines representing the true values.}
\label{SVAR psnpe posterior}
\end{center}
\end{figure}

The estimated marginal posterior plots are displayed in Figure \ref{SVAR psnpe posterior}, where the black solid lines represent the true parameter values. It is evident that our PNPE method (in a single round) produces a substantially sharper approximation of the posterior compared to that of SNPE. 

\begin{figure}[H]
\begin{center}
\centerline{\includegraphics[width=1\columnwidth]{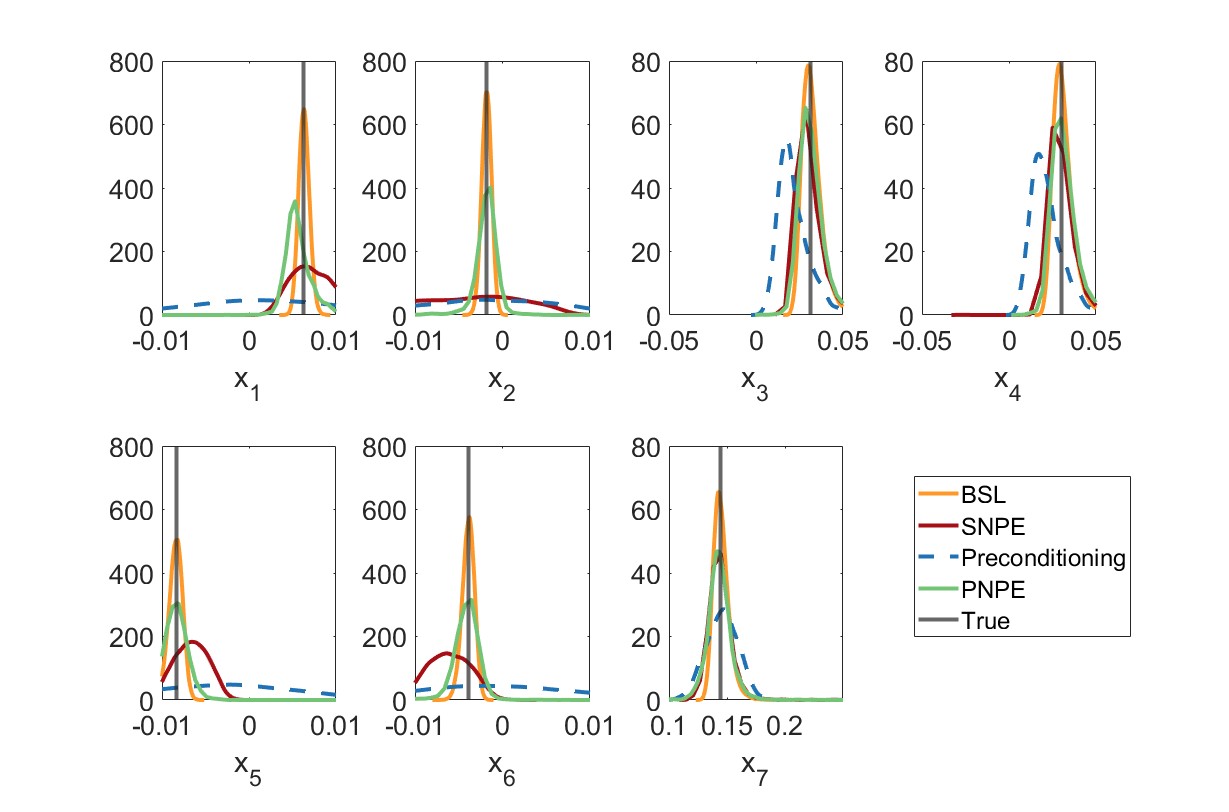}}
\caption{\textbf{Performance on SVAR model.} Comparison of posterior predictive distributions of the summary statistics of observation datasets between BSL (orange), SNPE (red), preconditioning step (blue dash) and PNPE (green solid), with black dashed lines representing the true values.}
\label{SVAR psnpe posterior predictive}
\end{center}
\end{figure}

It is evident that even a short run of the ABC algorithm gives a reasonable posterior approximation to train an unconditional normalizing flow as the unconditional density estimator, which then generates samples for NPE training. Figures \ref{SVAR psnpe posterior} and \ref{SVAR psnpe posterior predictive} show that the improved parameter posteriors leads to more accurate posterior predictive distributions of the summaries compared to SNPE. This indicates that with a good starting point, NPE can further improve accuracy.

\section{Further Experiments}\label{experiments}
We present two additional examples, including our motivating example, where SNPE, perhaps surprisingly, does not produce highly accurate posterior distributions. To fairly compare our method with vanilla SNPE and potentially TSNPE, we run the ABC algorithm 10 times and compute the average total number of simulations that the ABC algorithm requires. We then use this same number of simulations for the initial NPE in both SNPE and TSNPE. For all experiments, we utilize the adaptive SMC ABC algorithm proposed by \citet{drovandi2011estimation} for the preconditioning part with tuning parameters $n = 1$k, $a=0.5$, $c=0.01$, and use an unconditional normalizing flows as the unconditional density estimator. For SNPE, we use $10$k samples for each round of training.

We find that even with a single round of PNPE, there can be a significant improvement in performance. The code is publicly available in: \hyperref[PNPE]{https://github.com/john-wang1015/PNPE}

\subsection{High-dimensional SVAR model}\label{SVAR section}
To investigate how our method scales to higher dimensional problems, we take the illustrative SVAR example from before and consider $k=20$, which leads to 21 parameters. We detail the experimental settings in Section \ref{hdsvar setting}. To ensure a fair comparison, we run the SMC ABC algorithm ten times using a $10\%$ acceptance rate as the stopping criterion and calculate the average number of simulations it takes. We then use the same number of simulations, approximately $n_{\mathrm{ABC}} \approx 45k$, to train the initial NPE. Hence, the total number of simulations for both PNPE and SNPE is the same (55k in total). To stabilize the training, we apply the same clipping technique used in the previous low-dimensional case, which results in approximately 11\% of the training samples being removed in the initial round of NPE and about 1\% to 2\% in the second rounds. Starting from the third round, SNPE is unable to sample any parameter values from the neural networks due to a severe leakage issue \citep{deistler2022truncated}.

\begin{figure}[H]
\begin{center}
\centerline{\includegraphics[width=1.2\columnwidth]{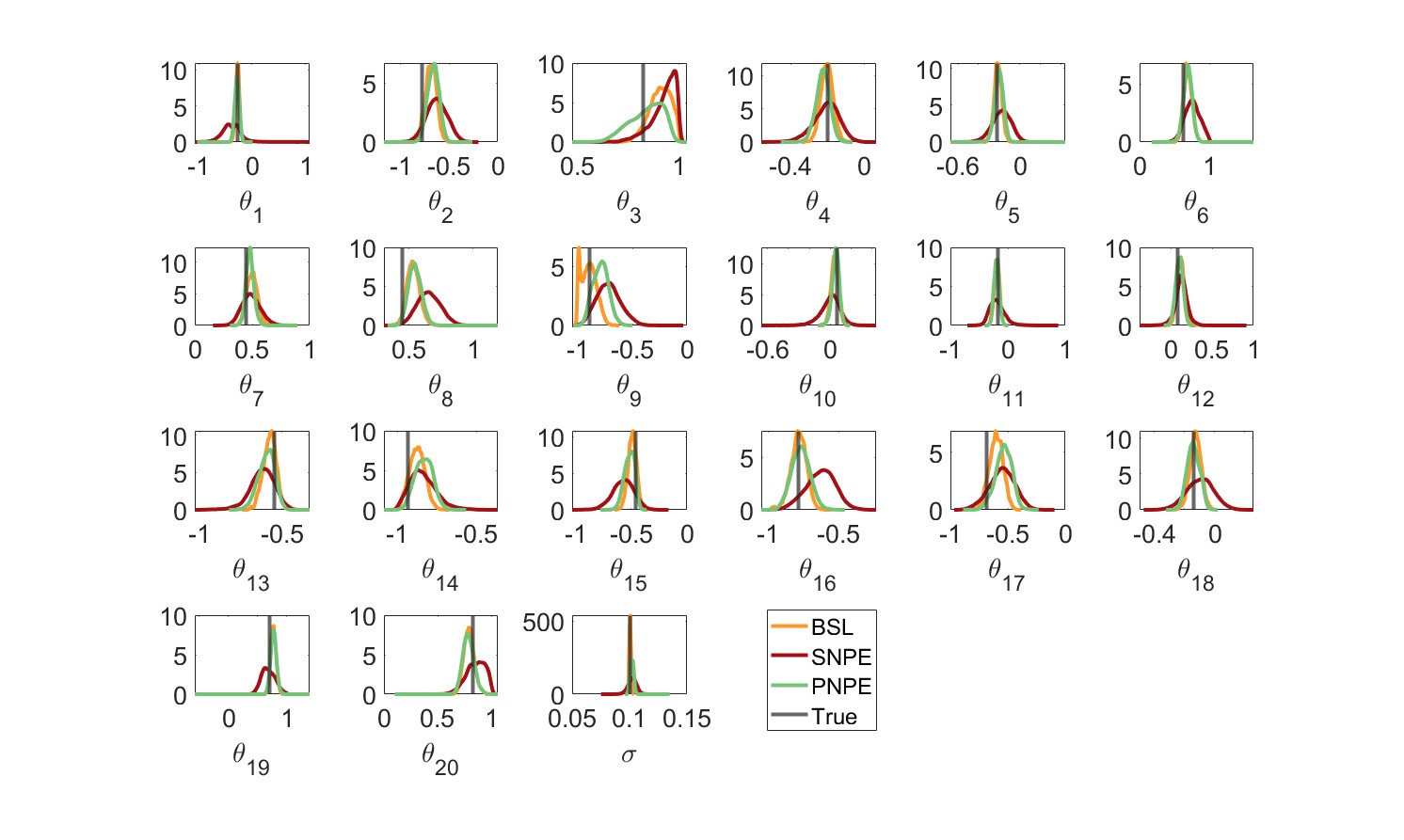}}
\caption{\textbf{Performance on SVAR model with 21 parameters.} Comparison of marginal posterior distributions between BSL (orange), SNPE (red) and PNPE (green), with black dashed lines representing the true values. The result of SNPE uses 2 rounds. }
\label{SVAR psnpe posterior hd}
\end{center}
\end{figure}

The estimated marginal posterior plots are displayed in Figure \ref{SVAR psnpe posterior hd}, where the black solid lines represent the true parameter values. It is evident that as the number of parameter dimensions increases, training the unconditional normalizing flows becomes more challenging. With well-trained unconditional normalizing flows, PNPE outperforms SNPE in high-dimensional cases (in this example, except for parameter $\theta_3$.). 

\begin{figure}[H]
\begin{center}
\centerline{\includegraphics[width=1.2\columnwidth]{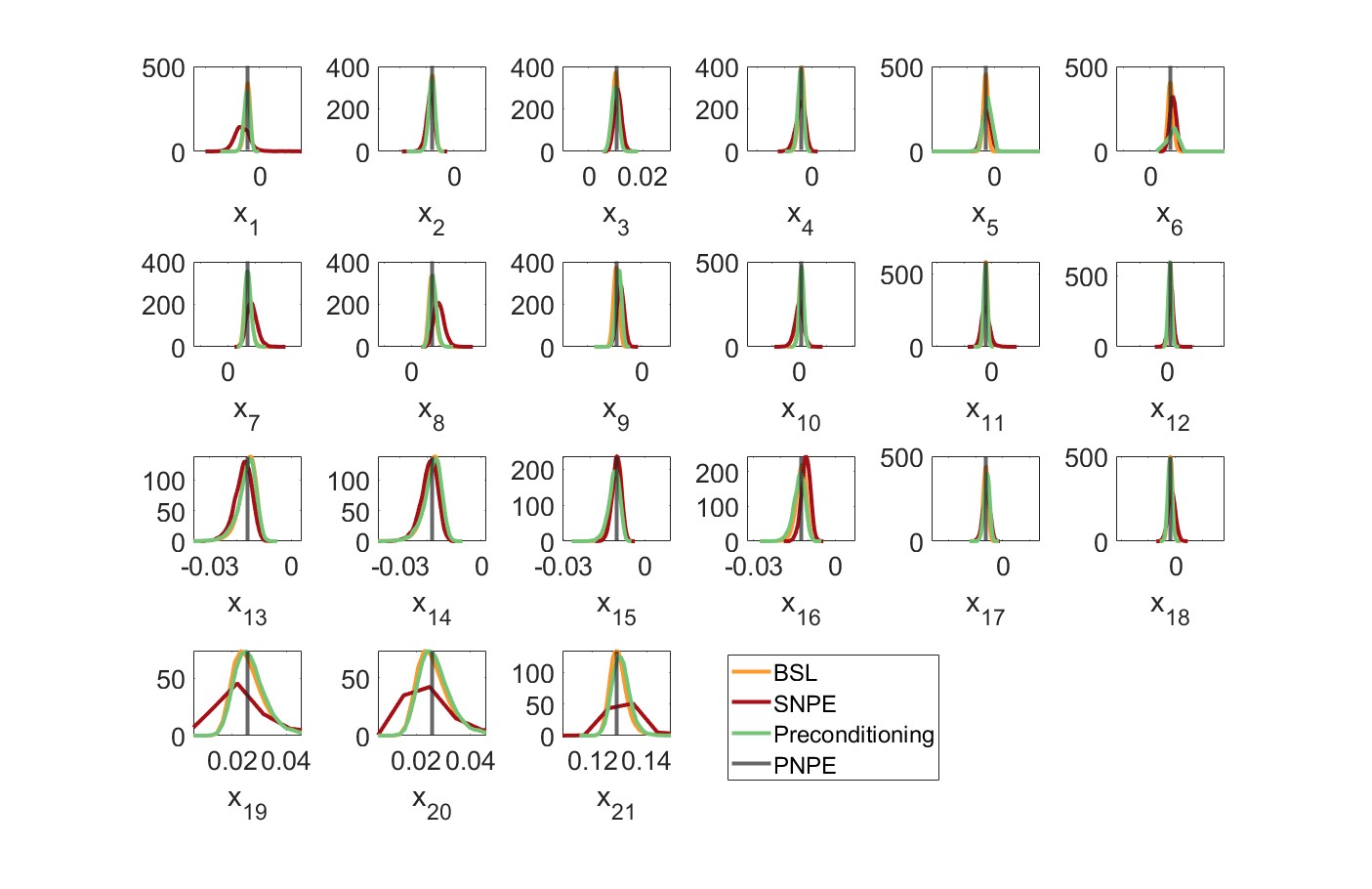}}
\caption{\textbf{Performance on SVAR model with 21 parameters.} Comparison of posterior predictive distributions of the summary statistics of observation datasets between BSL (orange), SNPE (red) and PNPE (green), with black dashed lines representing the true values. The result of SNPE uses 2 rounds.}
\label{SVAR psnpe posterior predictive hd}
\end{center}
\end{figure}



\subsection{Biphasic Voronoi cell-based model}\label{BVCBM}
Finally, we consider a challenging real-world problem in cancer biology: calibrating the biphasic Voronoi cell-based model (BVCBM) \cite{wang2022calibration} that models tumor growth. The model uses a parameter $\tau$ to divide the tumor into two growth phases. Here, the term `growth phase' refers to the different growth patterns of the tumor. There are four parameters that govern tumor growth during each phase, namely $(p_0, p_{\mathrm{psc}}, d_{\mathrm{max}}, g_{\mathrm{age}})$,  where $p_0$ and $p_{\mathrm{psc}}$ are the probability of cell proliferation and invasion, respectively, $d_{\mathrm{max}}$ is the maximum distance between cell and nutrient, and $g_{\mathrm{age}}$ is the time taken for a cell to be able to divide.  Thus there is nine parameters in total, four parameters each of two phases, and the parameter $\tau$ at which the growth phase changes. In this paper, we calibrate to two real-world pancreatic cancer datasets \cite{wade2019fabrication}, which describe tumor growth as time series data.  The datasets span 26 and 32 days, respectively, with measurements taken each day.  While the ground truth posteriors are unknown for those datasets, we compute posterior predictive distributions to assess if SNPE and PNPE can effectively calibrate the model to the data. 

We employ vague prior distributions for all parameters. Specifically, we use a Uniform distribution constrained between 1 and 24 hours $\times$ the number of days for $g_{\mathrm{age}}$ during both growth phases. Additionally, we use a Uniform distribution constrained between 2 and the number of days minus 1 for $\tau$. The prior distributions for the remaining parameters are detailed in Appendix \ref{BVCBM details}. 

As reported by \citet{wang2022calibration}, CPU times for model simulation range from 1.76 to 137.27 seconds per simulation when using samples from the prior distribution. This implies that 10k simulations for the first round of SNPE take approximately 2 hours. Consequently, the initial stages of SNPE are computationally expensive. In contrast, the ABC preconditioning step takes around 10-15 minutes. This is due to the fact that the longer simulation times tend to also lead to large discrepancies with the observed data, and such samples are quickly rejected by ABC. For the ABC part, around 18k and 16k simulations are used for the 26-day and 32-day pancreatic cancer datasets, respectively.

We observed a leakage problem occurring in rounds 8 and 5 for the 26-day and 32-day datasets, respectively. At round 10, the acceptance rate of SNPE for the 26-day measurement dataset is above 50$\%$, leading us to utilize rejection sampling for sample generation. While a 50$\%$ acceptance rate for rejection sampling is acceptable, it is less efficient compared to direct simulation from the trained conditional normalizing flows in the previous round. Hence, we employ TSNPE for this dataset. For the 32-day measurement dataset, only 0.000$\%$ of samples are accepted at round 8, making it computationally expensive. Consequently, we use TSNPE for this dataset.

\begin{figure}[h]
\begin{center}
\centerline{\includegraphics[width=1\columnwidth]{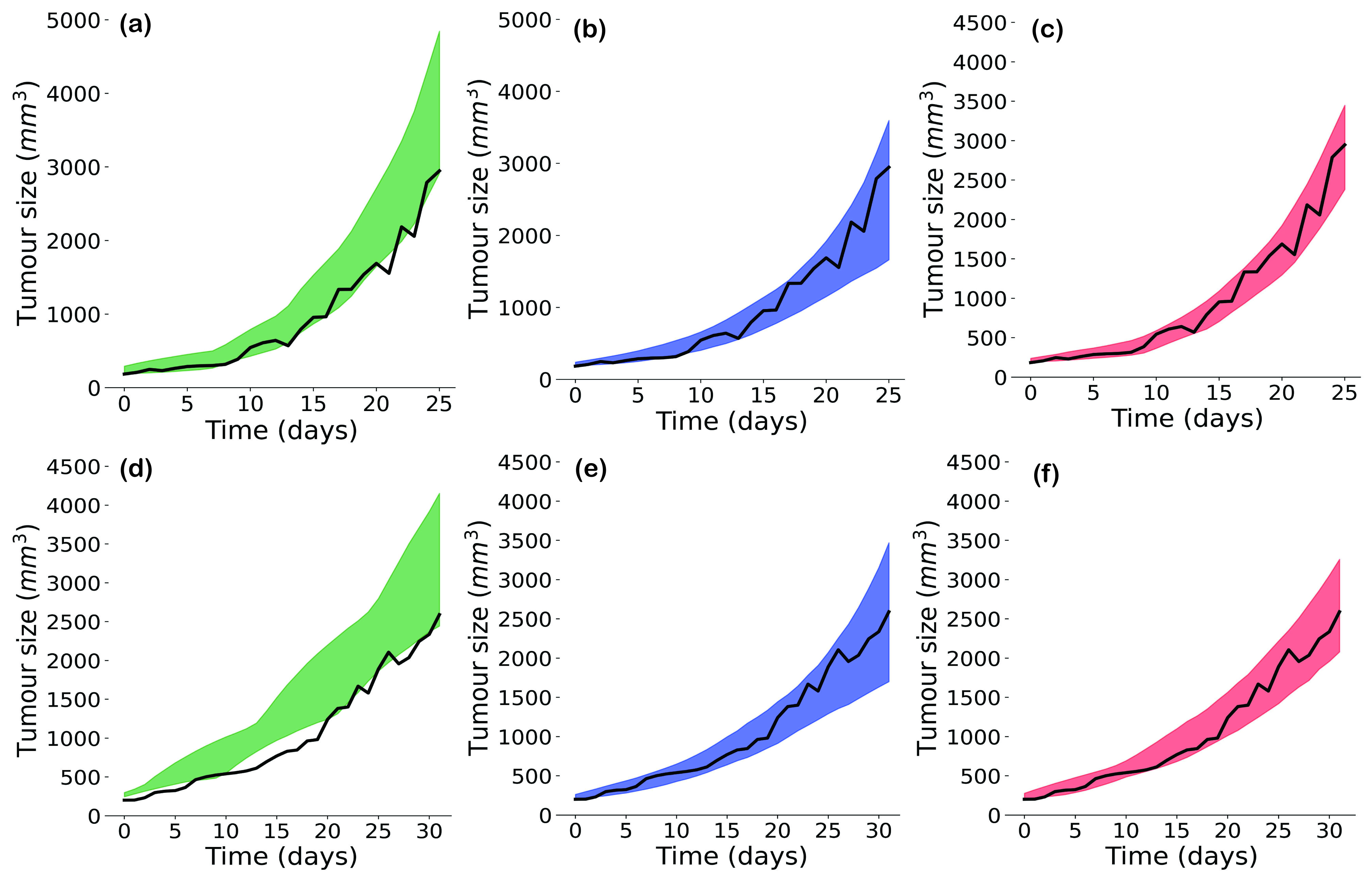}}
\caption{\textbf{Posterior predictive distributions for two pancreatic cancer datasets.} 90\% Posterior predictive interval plots for SNPE (\textbf{a}), preconditioning step (\textbf{b}) and PNPE (\textbf{c}) for the 26-day dataset; 90\% Posterior predictive interval plots for SNPE (\textbf{d}), preconditioning step (\textbf{e}) and PNPE (\textbf{f}) for the 32-day dataset.}
\label{BVCBM predictive}
\end{center}
\end{figure}

To estimate the posterior predictive distributions, we sample 1k parameter values from (T)SNPE and PNPE, using them to simulate datasets. We then plot these data in the form of credible intervals.  As a baseline, we show the prior predictive distributions in Appendix \ref{ferbvcbm}. The top and bottom rows of Figure \ref{BVCBM predictive} displays the posterior predictive distribution for the 26-day and 32-day datasets, respectively (the same plots but on the log scale are shown in Appendix \ref{ferbvcbm}). It is evident that both SNPE and TSNPE provide biased estimations, as the posterior predictive does not capture the observed data well.  The posterior predictive distribution for the preconditioning step (middle column) shows that the ABC step can capture the data reasonably well since the observed data lie within the 90$\%$ posterior predictive interval. Our method (third column) provides a better fit, as the variance of the posterior predictive distribution is tighter than that of the preconditioning step and still captures the observed data. This demonstrates that even one round of PNPE can perform more accurate estimations based on our results.

\section{Discussion}
We present a neural SBI method that is both simple and easy to deploy, designed to enhance the accuracy of SNPE methods. Our method, termed preconditioned neural posterior estimation (PNPE) and its sequential version, PSNPE, employs an ABC algorithm for the initial step. This algorithm is used to efficiently filter out poor regions of the parameter space. Additionally, we use the ABC posterior samples to train an unconditional density estimator $q_G$, enabling $q_G$ to serve as the initial proposal distribution for SNPE. The core concept is that an improved starting point can significantly enhance the accuracy of SNPE estimations.  Indeed, we obtained very good results with PNPE.

We showcase several examples where either SNPE failed to perform inference effectively, such as in the SVAR case, or produced biased results, as observed in the BVCBM. For the SVAR example, SNPE methods struggle due to the impact of low-quality samples from certain parameter space regions, adversely affecting the training process. The ABC method can efficiently eliminate these bad samples, thereby enhancing the training. For cases where SNPE results in biased estimations, our methods were effective at accurately fitting observed data (real data for BVCBM example). This is substantiated by our empirical results for the posterior predictive distribution discussed in experimental section.

Although our method demonstrates the capability to enhance estimation accuracy, it does have some limitations. Firstly, our method requires model simulations in the ABC preconditioning step, which may lead to greater computational demands in situations where SNPE methods perform well. However, by performing the preconditioning step, significantly fewer model simulations may be required in the SNPE part to achieve high accuracy.  In this paper we used SMC ABC for the preconditioning step, but we note that other ABC algorithms or SBI methods could be used.  We do not recycle the simulations performed in the ABC preconditioning step for the SNPE phase, but it could be possible to modify our method to exploit these model simulations.  Here we used the acceptance rate as the stopping criterion for the preconditioning step, but another option could be to check after each iteration of SMC ABC and stop the preconditioning if a suitable NCDE is found.  Secondly, the choice of an unconditional density estimator necessitates careful consideration. In scenarios involving low-dimensional parameter spaces, a kernel density estimator might be a preferable option compared to unconditional normalizing flows.

It is important to note that the preconditioning step of our method requires manual selection of the summary statistics or discrepancy function. However, this choice is not as critical as in a typical ABC application, since we aim to remove poor parts of the parameter space rather than achieve a highly accurate posterior approximation. In this paper we used the same summary statistics as in the preconditioning step for the subsequent rounds of NPE. However, we note it would be possible to use different summaries after the preconditioning step, and possibly use automated summary statistic selection methods \citep{fearnhead2012constructing,chen2023learning}.

In this paper we considered the well-specified scenario, where the model is either known to be correct or can provide a good fit to the data with a suitable choice of parameter values.  Standard neural SBI methods are known to potentially perform poorly under model misspecification \cite{bon2023being, cannon2022investigating, schmitt2021detecting}.  Our preconditioning method may be useful in the misspecified scenario, since ABC are known to perform reasonably well under model misspecification.  That is,  ABC still converges onto the pseudo-true parameter value \cite{frazier2020model}.  The preconditioning step could be followed by a robust neural SBI method such as \citet{kelly2023misspecification,huang2023learning,glockler2023adversarial,ward2022robust}.  We plan to explore this in future research.

Overall, PNPE employs a preconditioning step to focus on important parts of the parameter space, thereby creating a good starting point for training SNPE and enhancing estimation accuracy. We have empirically demonstrated that PNPE is capable of producing more accurate estimations in complex real-world problems.

\section*{Acknowledgement}
We thank the computational resources provided by QUT’s High Performance Computing and Research Support Group (HPC). Xiaoyu Wang, Ryan P. Kelly and Christopher Drovandi were supported by an Australian Research Council Future Fellowship (FT210100260).

\bibliography{main}

\begin{thebibliography}{51}
\providecommand{\natexlab}[1]{#1}
\providecommand{\url}[1]{\texttt{#1}}
\expandafter\ifx\csname urlstyle\endcsname\relax
  \providecommand{\doi}[1]{doi: #1}\else
  \providecommand{\doi}{doi: \begingroup \urlstyle{rm}\Url}\fi

\bibitem[Aylett-Bullock et~al.(2021)Aylett-Bullock, Cuesta-Lazaro, Quera-Bofarull, Icaza-Lizaola, Sedgewick, Truong, Curran, Elliott, Caulfield, Fong, et~al.]{aylett2021june}
J.~Aylett-Bullock, C.~Cuesta-Lazaro, A.~Quera-Bofarull, M.~Icaza-Lizaola, A.~Sedgewick, H.~Truong, A.~Curran, E.~Elliott, T.~Caulfield, K.~Fong, et~al.
\newblock June: open-source individual-based epidemiology simulation.
\newblock \emph{Royal Society {O}pen {S}cience}, 8\penalty0 (7):\penalty0 210506, 2021.

\bibitem[Beaumont(2019)]{beaumont2019approximate}
M.~A. Beaumont.
\newblock Approximate {B}ayesian computation.
\newblock \emph{Annual {R}eview of {S}tatistics and {I}ts {A}pplication}, 6:\penalty0 379--403, 2019.

\bibitem[Beaumont et~al.(2009)Beaumont, Cornuet, Marin, and Robert]{beaumont2009adaptive}
M.~A. Beaumont, J.-M. Cornuet, J.-M. Marin, and C.~P. Robert.
\newblock Adaptive approximate {B}ayesian computation.
\newblock \emph{Biometrika}, 96\penalty0 (4):\penalty0 983--990, 2009.

\bibitem[Biau et~al.(2015)Biau, C{\'e}rou, and Guyader]{biau2015new}
G.~Biau, F.~C{\'e}rou, and A.~Guyader.
\newblock New insights into approximate {B}ayesian computation.
\newblock In \emph{Annales de l'{IHP} {P}robabilit{\'e}s et {S}tatistiques}, volume~51, pages 376--403, 2015.

\bibitem[Bingham et~al.(2019)Bingham, Chen, Jankowiak, Obermeyer, Pradhan, Karaletsos, Singh, Szerlip, Horsfall, and Goodman]{bingham2019pyro}
E.~Bingham, J.~P. Chen, M.~Jankowiak, F.~Obermeyer, N.~Pradhan, T.~Karaletsos, R.~Singh, P.~Szerlip, P.~Horsfall, and N.~D. Goodman.
\newblock Pyro: Deep universal probabilistic programming.
\newblock \emph{The {J}ournal of {M}achine {L}earning {R}esearch}, 20\penalty0 (1):\penalty0 973--978, 2019.

\bibitem[Blum(2010)]{blum2010approximate}
M.~G. Blum.
\newblock Approximate {B}ayesian computation: a nonparametric perspective.
\newblock \emph{{Journal of the American Statistical Association}}, 105\penalty0 (491):\penalty0 1178--1187, 2010.

\bibitem[Bon et~al.(2023)Bon, Bretherton, Buchhorn, Cramb, Drovandi, Hassan, Jenner, Mayfield, McGree, Mengersen, et~al.]{bon2023being}
J.~J. Bon, A.~Bretherton, K.~Buchhorn, S.~Cramb, C.~Drovandi, C.~Hassan, A.~L. Jenner, H.~J. Mayfield, J.~M. McGree, K.~Mengersen, et~al.
\newblock Being {B}ayesian in the 2020s: opportunities and challenges in the practice of modern applied {B}ayesian statistics.
\newblock \emph{Philosophical {T}ransactions of the {R}oyal {S}ociety {A}}, 381\penalty0 (2247):\penalty0 20220156, 2023.

\bibitem[Cannon et~al.(2022)Cannon, Ward, and Schmon]{cannon2022investigating}
P.~Cannon, D.~Ward, and S.~M. Schmon.
\newblock Investigating the impact of model misspecification in neural simulation-based inference.
\newblock \emph{arXiv preprint arXiv:2209.01845}, 2022.

\bibitem[Chen et~al.(2023)Chen, Gutmann, and Weller]{chen2023learning}
Y.~Chen, M.~U. Gutmann, and A.~Weller.
\newblock Is learning summary statistics necessary for likelihood-free inference?
\newblock In \emph{{International Conference on Machine Learning}}, pages 4529--4544. PMLR, 2023.

\bibitem[Cranmer et~al.(2020)Cranmer, Brehmer, and Louppe]{cranmer2020frontier}
K.~Cranmer, J.~Brehmer, and G.~Louppe.
\newblock The frontier of simulation-based inference.
\newblock \emph{Proceedings of the National Academy of Sciences}, 117\penalty0 (48):\penalty0 30055--30062, 2020.

\bibitem[Csill{\'e}ry et~al.(2010)Csill{\'e}ry, Blum, Gaggiotti, and Fran{\c{c}}ois]{csillery2010approximate}
K.~Csill{\'e}ry, M.~G. Blum, O.~E. Gaggiotti, and O.~Fran{\c{c}}ois.
\newblock Approximate {B}ayesian computation ({ABC}) in practice.
\newblock \emph{Trends in {E}cology \& {E}volution}, 25\penalty0 (7):\penalty0 410--418, 2010.

\bibitem[Dax et~al.(2021)Dax, Green, Gair, Macke, Buonanno, and Sch{\"o}lkopf]{dax2021real}
M.~Dax, S.~R. Green, J.~Gair, J.~H. Macke, A.~Buonanno, and B.~Sch{\"o}lkopf.
\newblock Real-time gravitational wave science with neural posterior estimation.
\newblock \emph{Physical {R}eview {L}etters}, 127\penalty0 (24):\penalty0 241103, 2021.

\bibitem[de~Santi et~al.(2023)de~Santi, Villaescusa-Navarro, Abramo, Shao, Perez, Castro, Ni, Lovell, Hernandez-Martinez, Marinacci, et~al.]{de2023field}
N.~S. de~Santi, F.~Villaescusa-Navarro, L.~R. Abramo, H.~Shao, L.~A. Perez, T.~Castro, Y.~Ni, C.~C. Lovell, E.~Hernandez-Martinez, F.~Marinacci, et~al.
\newblock Field-level simulation-based inference with {G}alaxy catalogs: the impact of systematic effects.
\newblock \emph{arXiv preprint arXiv:2310.15234}, 2023.

\bibitem[Deistler et~al.(2022)Deistler, Goncalves, and Macke]{deistler2022truncated}
M.~Deistler, P.~J. Goncalves, and J.~H. Macke.
\newblock Truncated proposals for scalable and hassle-free simulation-based inference.
\newblock \emph{Advances in {N}eural {I}nformation {P}rocessing {S}ystems}, 35:\penalty0 23135--23149, 2022.

\bibitem[Dolatabadi et~al.(2020)Dolatabadi, Erfani, and Leckie]{dolatabadi2020invertible}
H.~M. Dolatabadi, S.~Erfani, and C.~Leckie.
\newblock Invertible generative modeling using linear rational splines.
\newblock In \emph{International Conference on Artificial Intelligence and Statistics}, pages 4236--4246. PMLR, 2020.

\bibitem[Drovandi et~al.(2023)Drovandi, Nott, and Frazier]{drovandi2023improving}
C.~Drovandi, D.~J. Nott, and D.~T. Frazier.
\newblock Improving the accuracy of marginal approximations in likelihood-free inference via localisation.
\newblock \emph{Journal of {C}omputational and {G}raphical {S}tatistics}, pages 1--19, 2023.

\bibitem[Drovandi and Pettitt(2011)]{drovandi2011estimation}
C.~C. Drovandi and A.~N. Pettitt.
\newblock Estimation of parameters for macroparasite population evolution using approximate {B}ayesian computation.
\newblock \emph{Biometrics}, 67\penalty0 (1):\penalty0 225--233, 2011.

\bibitem[Durkan et~al.(2019)Durkan, Bekasov, Murray, and Papamakarios]{durkan2019neural}
C.~Durkan, A.~Bekasov, I.~Murray, and G.~Papamakarios.
\newblock Neural spline flows.
\newblock \emph{Advances in {N}eural {I}nformation {P}rocessing {S}ystems}, 32, 2019.

\bibitem[Fearnhead and Prangle(2012)]{fearnhead2012constructing}
P.~Fearnhead and D.~Prangle.
\newblock Constructing summary statistics for approximate {B}ayesian computation: semi-automatic approximate {B}ayesian computation.
\newblock \emph{Journal of the {R}oyal {S}tatistical {S}ociety {S}eries {B}: {S}tatistical {M}ethodology}, 74\penalty0 (3):\penalty0 419--474, 2012.

\bibitem[Fengler et~al.(2021)Fengler, Govindarajan, Chen, and Frank]{fengler2021likelihood}
A.~Fengler, L.~N. Govindarajan, T.~Chen, and M.~J. Frank.
\newblock Likelihood approximation networks ({LANs}) for fast inference of simulation models in cognitive neuroscience.
\newblock \emph{Elife}, 10:\penalty0 e65074, 2021.

\bibitem[Frazier et~al.(2020)Frazier, Robert, and Rousseau]{frazier2020model}
D.~T. Frazier, C.~P. Robert, and J.~Rousseau.
\newblock Model misspecification in approximate {B}ayesian computation: consequences and diagnostics.
\newblock \emph{Journal of the {R}oyal {S}tatistical {S}ociety {S}eries {B}: {S}tatistical {M}ethodology}, 82\penalty0 (2):\penalty0 421--444, 2020.

\bibitem[Gloeckler et~al.(2023)Gloeckler, Deistler, and Macke]{glockler2023adversarial}
M.~Gloeckler, M.~Deistler, and J.~H. Macke.
\newblock Adversarial robustness of amortized {B}ayesian inference.
\newblock pages 11493--11524, 2023.

\bibitem[Greenberg et~al.(2019)Greenberg, Nonnenmacher, and Macke]{greenberg2019automatic}
D.~Greenberg, M.~Nonnenmacher, and J.~Macke.
\newblock Automatic posterior transformation for likelihood-free inference.
\newblock In \emph{International {C}onference on {M}achine {L}earning}, pages 2404--2414. PMLR, 2019.

\bibitem[Huang et~al.(2024)Huang, Bharti, Souza, Acerbi, and Kaski]{huang2023learning}
D.~Huang, A.~Bharti, A.~Souza, L.~Acerbi, and S.~Kaski.
\newblock Learning robust statistics for simulation-based inference under model misspecification.
\newblock \emph{{Advances in Neural Information Processing Systems}}, 36, 2024.

\bibitem[Jenner et~al.(2020)Jenner, Frascoli, Coster, and Kim]{jenner2020enhancing}
A.~L. Jenner, F.~Frascoli, A.~C. Coster, and P.~S. Kim.
\newblock Enhancing oncolytic virotherapy: {O}bservations from a {V}oronoi {C}ell-{B}ased model.
\newblock \emph{Journal of {T}heoretical {B}iology}, 485:\penalty0 110052, 2020.

\bibitem[Kelly et~al.(2023)Kelly, Nott, Frazier, Warne, and Drovandi]{kelly2023misspecification}
R.~P. Kelly, D.~J. Nott, D.~T. Frazier, D.~J. Warne, and C.~Drovandi.
\newblock Misspecification-robust sequential neural likelihood.
\newblock \emph{arXiv preprint arXiv:2301.13368}, 2023.

\bibitem[Kim et~al.(2011)Kim, Sohn, Choi, Jung, Kim, Haam, and Yun]{kim2011active}
P.-H. Kim, J.-H. Sohn, J.-W. Choi, Y.~Jung, S.~W. Kim, S.~Haam, and C.-O. Yun.
\newblock Active targeting and safety profile of {PEG}-modified adenovirus conjugated with herceptin.
\newblock \emph{Biomaterials}, 32\penalty0 (9):\penalty0 2314--2326, 2011.

\bibitem[Lintusaari et~al.(2017)Lintusaari, Gutmann, Dutta, Kaski, and Corander]{lintusaari2017fundamentals}
J.~Lintusaari, M.~U. Gutmann, R.~Dutta, S.~Kaski, and J.~Corander.
\newblock Fundamentals and recent developments in approximate {B}ayesian computation.
\newblock \emph{Systematic {B}iology}, 66\penalty0 (1):\penalty0 e66--e82, 2017.

\bibitem[Lueckmann et~al.(2017)Lueckmann, Goncalves, Bassetto, {\"O}cal, Nonnenmacher, and Macke]{lueckmann2017flexible}
J.-M. Lueckmann, P.~J. Goncalves, G.~Bassetto, K.~{\"O}cal, M.~Nonnenmacher, and J.~H. Macke.
\newblock Flexible statistical inference for mechanistic models of neural dynamics.
\newblock \emph{Advances in {N}eural {I}nformation {P}rocessing {S}ystems}, 30, 2017.

\bibitem[Lueckmann et~al.(2021)Lueckmann, Boelts, Greenberg, Goncalves, and Macke]{lueckmann2021benchmarking}
J.-M. Lueckmann, J.~Boelts, D.~Greenberg, P.~Goncalves, and J.~Macke.
\newblock Benchmarking simulation-based inference.
\newblock In \emph{International {C}onference on {A}rtificial {I}ntelligence and {S}tatistics}, pages 343--351. PMLR, 2021.

\bibitem[Meineke et~al.(2001)Meineke, Potten, and Loeffler]{meineke2001cell}
F.~A. Meineke, C.~S. Potten, and M.~Loeffler.
\newblock Cell migration and organization in the intestinal crypt using a lattice-free model.
\newblock \emph{Cell {P}roliferation}, 34\penalty0 (4):\penalty0 253--266, 2001.

\bibitem[Mishra-Sharma(2022)]{mishra2022inferring}
S.~Mishra-Sharma.
\newblock Inferring dark matter substructure with astrometric lensing beyond the power spectrum.
\newblock \emph{Machine Learning: Science and Technology}, 3\penalty0 (1):\penalty0 01LT03, 2022.

\bibitem[Papamakarios and Murray(2016)]{papamakarios2016fast}
G.~Papamakarios and I.~Murray.
\newblock Fast $\varepsilon$-free inference of simulation models with {B}ayesian conditional density estimation.
\newblock \emph{Advances in {N}eural {I}nformation {P}rocessing {S}ystems}, 29, 2016.

\bibitem[Papamakarios et~al.(2017)Papamakarios, Pavlakou, and Murray]{papamakarios2017masked}
G.~Papamakarios, T.~Pavlakou, and I.~Murray.
\newblock Masked autoregressive flow for density estimation.
\newblock \emph{Advances in {N}eural {I}nformation {P}rocessing {S}ystems}, 30, 2017.

\bibitem[Papamakarios et~al.(2021)Papamakarios, Nalisnick, Rezende, Mohamed, and Lakshminarayanan]{papamakarios2021normalizing}
G.~Papamakarios, E.~Nalisnick, D.~J. Rezende, S.~Mohamed, and B.~Lakshminarayanan.
\newblock Normalizing flows for probabilistic modeling and inference.
\newblock \emph{The {J}ournal of {M}achine {L}earning {R}esearch}, 22\penalty0 (1):\penalty0 2617--2680, 2021.

\bibitem[Pedregosa et~al.(2011)Pedregosa, Varoquaux, Gramfort, Michel, Thirion, Grisel, Blondel, Prettenhofer, Weiss, Dubourg, et~al.]{pedregosa2011scikit}
F.~Pedregosa, G.~Varoquaux, A.~Gramfort, V.~Michel, B.~Thirion, O.~Grisel, M.~Blondel, P.~Prettenhofer, R.~Weiss, V.~Dubourg, et~al.
\newblock Scikit-learn: {M}achine learning in {P}ython.
\newblock \emph{the {J}ournal of {M}achine {L}earning {R}esearch}, 12:\penalty0 2825--2830, 2011.

\bibitem[Price et~al.(2018)Price, Drovandi, Lee, and Nott]{price2018bayesian}
L.~F. Price, C.~C. Drovandi, A.~Lee, and D.~J. Nott.
\newblock Bayesian synthetic likelihood.
\newblock \emph{{Journal of Computational and Graphical Statistics}}, 27\penalty0 (1):\penalty0 1--11, 2018.

\bibitem[Rezende and Mohamed(2015)]{rezende2015variational}
D.~Rezende and S.~Mohamed.
\newblock Variational inference with normalizing flows.
\newblock In \emph{International {C}onference on {M}achine {L}earning}, pages 1530--1538. PMLR, 2015.

\bibitem[Schmitt et~al.(2023)Schmitt, B{\"u}rkner, K{\"o}the, and Radev]{schmitt2021detecting}
M.~Schmitt, P.-C. B{\"u}rkner, U.~K{\"o}the, and S.~T. Radev.
\newblock Detecting model misspecification in amortized {B}ayesian inference with neural networks.
\newblock pages 541--557, 2023.

\bibitem[Shih et~al.(2023)Shih, Freytsis, Taylor, Dror, and Smyth]{shih2023fast}
D.~Shih, M.~Freytsis, S.~R. Taylor, J.~A. Dror, and N.~Smyth.
\newblock Fast parameter inference on pulsar timing arrays with normalizing flows.
\newblock \emph{arXiv preprint arXiv:2310.12209}, 2023.

\bibitem[Sisson et~al.(2007)Sisson, Fan, and Tanaka]{sisson2007sequential}
S.~A. Sisson, Y.~Fan, and M.~M. Tanaka.
\newblock Sequential {M}onte {C}arlo without likelihoods.
\newblock \emph{Proceedings of the {N}ational {A}cademy of {S}ciences}, 104\penalty0 (6):\penalty0 1760--1765, 2007.

\bibitem[Sisson et~al.(2018)Sisson, Fan, and Beaumont]{sisson2018handbook}
S.~A. Sisson, Y.~Fan, and M.~Beaumont.
\newblock \emph{Handbook of approximate Bayesian computation}.
\newblock CRC Press, 2018.

\bibitem[Tejero-Cantero et~al.(2020)Tejero-Cantero, Boelts, Deistler, Lueckmann, Durkan, Gon{\c{c}}alves, Greenberg, and Macke]{tejero2020sbi}
A.~Tejero-Cantero, J.~Boelts, M.~Deistler, J.-M. Lueckmann, C.~Durkan, P.~J. Gon{\c{c}}alves, D.~S. Greenberg, and J.~H. Macke.
\newblock {SBI}: A toolkit for simulation-based inference.
\newblock \emph{{Journal of Open Source Software}}, 5\penalty0 (52):\penalty0 2505, 2020.

\bibitem[Thomas et~al.(2020)Thomas, Pesonen, S{\'a}-Le{\~a}o, de~Lencastre, Kaski, and Corander]{thomas2020split}
O.~Thomas, H.~Pesonen, R.~S{\'a}-Le{\~a}o, H.~de~Lencastre, S.~Kaski, and J.~Corander.
\newblock Split-{BOLFI} for for misspecification-robust likelihood free inference in high dimensions.
\newblock \emph{arXiv preprint arXiv:2002.09377}, 2020.

\bibitem[Turner and Van~Zandt(2012)]{turner2012tutorial}
B.~M. Turner and T.~Van~Zandt.
\newblock A tutorial on approximate {B}ayesian computation.
\newblock \emph{Journal of Mathematical Psychology}, 56\penalty0 (2):\penalty0 69--85, 2012.

\bibitem[Wade(2019)]{wade2019fabrication}
S.~J. Wade.
\newblock Fabrication and preclinical assessment of drug eluting wet spun fibres for pancreatic cancer treatment.
\newblock 2019.

\bibitem[Wang et~al.(2024)Wang, Jenner, Salomone, Warne, and Drovandi]{wang2022calibration}
X.~Wang, A.~L. Jenner, R.~Salomone, D.~J. Warne, and C.~Drovandi.
\newblock Calibration of agent based models for monophasic and biphasic tumour growth using approximate {B}ayesian computation.
\newblock \emph{{Journal of Mathematical Biology}}, 88\penalty0 (3):\penalty0 28, 2024.

\bibitem[Ward et~al.(2022)Ward, Cannon, Beaumont, Fasiolo, and Schmon]{ward2022robust}
D.~Ward, P.~Cannon, M.~Beaumont, M.~Fasiolo, and S.~Schmon.
\newblock Robust neural posterior estimation and statistical model criticism.
\newblock \emph{Advances in {N}eural {I}nformation {P}rocessing {S}ystems}, 35:\penalty0 33845--33859, 2022.

\bibitem[Wehenkel et~al.(2023)Wehenkel, Behrmann, Miller, Sapiro, Sener, Cuturi, and Jacobsen]{wehenkel2023simulation}
A.~Wehenkel, J.~Behrmann, A.~C. Miller, G.~Sapiro, O.~Sener, M.~Cuturi, and J.-H. Jacobsen.
\newblock Simulation-based inference for cardiovascular models.
\newblock \emph{arXiv preprint arXiv:2307.13918}, 2023.

\bibitem[West et~al.(2021)West, Berthouze, Farmer, Cagnan, and Litvak]{west2021inference}
T.~O. West, L.~Berthouze, S.~F. Farmer, H.~Cagnan, and V.~Litvak.
\newblock Inference of brain networks with approximate {B}ayesian computation--assessing face validity with an example application in {P}arkinsonism.
\newblock \emph{Neuroimage}, 236:\penalty0 118020, 2021.

\bibitem[Winkler et~al.(2019)Winkler, Worrall, Hoogeboom, and Welling]{winkler2019learning}
C.~Winkler, D.~Worrall, E.~Hoogeboom, and M.~Welling.
\newblock Learning likelihoods with conditional normalizing flows.
\newblock \emph{arXiv preprint arXiv:1912.00042}, 2019.

\end{thebibliography}

\newpage
\appendix
\section{Further background for SMC ABC}\label{smc-abc details}

We provide a detailed description of the adaptive SMC ABC algorithm we used in this paper and provide pseudocode in Algorithm \ref{adaptive SMC ABC} for reference. This algorithm starts by drawing $N$ independent samples from the prior distribution $p({\theta})$, represented as $\{\theta_i\}_{i=1}^N$. For each sample ${\theta}_i$ (known as a particle), the algorithm simulates a dataset $x_i$ from the stochastic model and calculates the corresponding discrepancy $\rho_i = \rho(x_i, x_o)$, resulting in the pair set $\{\theta_i,\rho_i\}_{i=1}^N$. These pairs are then arranged in order of increasing discrepancy such that $\rho_1<\rho_2<\cdots<\rho_N$. The first tolerance threshold, $\epsilon_1$, is set as the largest discrepancy, $\rho_N$. To move through the target distributions, the algorithm adjusts the tolerance dynamically. The next tolerance, $\epsilon_{t}$, is set as $\rho_{N-N_a}$, where $N_a = \lfloor Na \rfloor$, and $a$ is a tuning parameter. Essentially, in each step, the algorithm discards the top $a\times100\%$ of particles with the highest discrepancies. After discarding these particles, only $N-N_a$ particles remain. To replenish the set back to $N$ particles, the algorithm resamples $N_a$ times from the `alive' particles, copying both the parameter and discrepancy values. This process, however, leads to duplicates in the particle set. To add variety to the set, the algorithm applies an MCMC ABC kernel to each resampled particle. The parameters for the MCMC proposal distribution $q_{t}(\cdot{\mid}\cdot)$ are derived from the current particle set. For instance, if using a multivariate normal random walk proposal, its covariance $\Sigma_t$ is based on the particle set's sample covariance. The acceptance of a proposed parameter (assuming a symmetric proposal) and simulated data is determined by the equation:
\begin{equation}
p_t = \mathrm{min}\left(1,\frac{p(\tilde{{\theta}})}{p({\theta})} \mathbb{I}(\rho({x_o},\tilde{{x}}) < \epsilon_t) \right),
\end{equation}
where $\tilde{{\theta}} \sim q(\cdot|\theta)$ and $\tilde{{x}} \sim p(\cdot|\tilde{{\theta}})$ are proposed parameter values and dataset, respectively.  However, proposals may be rejected, leaving some particles unchanged. To address this, the algorithm performs $R_t$ iterations of the MCMC kernel on each particle, where $R_t = \left\lceil \frac{\log(c)}{\log(1-p_t^{\mathrm{acc}})}\right\rceil$,  where $c$ is a tuning parameter of the algorithm that can be interpreted as the probability that a particle is not moved in the $R_t$ iterations. The acceptance probability $p_t^{\mathrm{acc}}$ is estimated from trial MCMC ABC iterations and used to compute $R_t$ for the next set of MCMC ABC iterations. For this adaptive SMC ABC algorithm, two stopping rules can be used. The first stopping rule halts the ABC algorithm when the maximum discrepancy is below a set tolerance, $\epsilon_T$. The second stopping rule terminates the algorithm when the MCMC acceptance probability $p_t^{\mathrm{acc}}$ falls below a predefined threshold $ p_{\mathrm{acc}} $. Here we choose the second rule, and since we only require a short run of ABC, we set $p_{\mathrm{acc}}$ to be higher than what is typically used in an ABC analysis.

\newpage
\begin{algorithm}[H]
   \caption{Adaptive SMC ABC}
   \label{adaptive SMC ABC}
\begin{algorithmic}
    \State {\bfseries Input:} The observed data $x_o$, the stochastic model $p({x}{\mid}{\theta})$, distance function $\rho(\cdot,\cdot)$, prior distribution $p({\theta})$, number of particles $N$, tuning parameters $a$ and $c$ for adaptive selection of discrepancy thresholds and selecting the number of MCMC iterations in the move steps, target tolerance $\epsilon_T$, initial number of trial MCMC iterations $S_{\mathrm{init}}$, minimum acceptable MCMC acceptance rate $p_{\mathrm{min}}$
    \For{$i=1,\dots,N$}
    \State Simulate $x_i \sim p(x|\theta_i)$ where $\theta_i \sim p(\theta)$
    \State Compute $\rho_i = \rho(x_o,x_i)$
    \EndFor
    \State Sort ${\{\theta_i\}}_{i=1}^N$ by ${\{\rho_i\}}_{i=1}^N$ such that $\rho_1 \leq \rho_2 \leq \dots \leq \rho_N$
    \State Set $N_{a} = \lfloor aN \rfloor, \quad t = 2, \quad$ $\epsilon_t = \rho_{N-N_a},\quad$ $\epsilon_{1} = \rho_N, \quad S_t = S_{\mathrm{init}}, \quad \tilde{p}_t^{\mathrm{acc}} = 1$
    \While{$\epsilon_{t-1} > \epsilon_T$ \textbf{or} $\tilde{p}_t^{\mathrm{acc}} > p_{\mathrm{min}}$}
    \State Compute $\Sigma$ as the sample covariance matrix of $\{{\theta}_{i}\}_{i=1}^{N-N_a}$
    \State Generate $\{{\theta}_{i}\}_{i = N-N_a+1}^N$ by resampling from $\{{\theta}_{i}\}_{i=1}^{N-N_a}$ with replacement 
    \For{$i = N - N_a+1,\dots,N$}
        \For{$j = 1,\dots,S_t$}
            \State Simulate $\tilde{{x}} \sim p({x}{\mid}\tilde{{\theta}})$ based on proposal $\tilde{{\theta}} \sim \mathcal{N}({{\theta}_i},\Sigma)$
            \State Compute $\tilde{\rho} = \rho({x_o},\tilde{{x}})$
            \State Compute $p_t^{i,j} = \mathrm{min}\left(1,\frac{p(\tilde{{\theta}})}{p({\theta})} \mathbb{I}(\tilde{\rho} < \epsilon_t)   \right).$
            \State With probability $p_t^{i,j}$, set ${\theta}_i = \tilde{{\theta}}$ and $\rho_i = \tilde{\rho}$; otherwise, retain the current values of ${\theta}_i$ and $\rho_i$
        \EndFor
    \EndFor
    \State ${\tilde{p}}_t = \sum_{i=N - N_a+1}^N \sum_{j=1}^{S_t} p_t^{i,j}/\big(S_t(N-N_{a})\big)$
    \State $R_t = \lceil \log(c)/\big(1 + \log(1-\tilde{p}_t)\big)\rceil$
    \For{$i = N - N_a+1,\dots,N$}
        \For{$j = R_t-S_t,\dots,R_t$}
            \State Simulate $\tilde{{x}} \sim p({x}{\mid}\tilde{{\theta}})$ based on proposal $\tilde{{\theta}} \sim \mathcal{N}({{\theta}_i},\Sigma)$
            \State Compute $\tilde{\rho} = \rho({x_o},\tilde{{x}})$
            \State Compute $p_t^{i,j} = \mathrm{min}\left(1,\frac{p(\tilde{{\theta}})}{p({\theta})} \mathbb{I}(\tilde{\rho} < \epsilon_t)   \right).$
            \State With probability $p_t^{i,j}$, set ${\theta}_i = \tilde{{\theta}}$ and $\rho_i = \tilde{\rho}$; otherwise, retain the current values of ${\theta}_i$ and $\rho_i$
        \EndFor
    \EndFor
    \State $\tilde{p}_t^{\mathrm{acc}} = \sum_{i=N-N_a+1}^N \sum_{j=1}^{R_t}p_t^{i,j}/\big(R_t(N-N_{a})\big)$
    \State $S_{t+1} = \lceil R_t/2\rceil$
    \State Sort $\{{\theta}_i\}_{i=1}^N$ by $\{\rho_{i}\}_{i=1}^N$ such that $\rho_{1} \leq \rho_{2} \leq \cdots \leq \rho_{N}$
    \State Set $\epsilon_{t+1} = \rho_{N-N_a}$, $\quad \epsilon_{t} = \rho_N$
    \State $t = t + 1$
    \EndWhile
    \State \textbf{return} Samples ${\{\theta_i\}}_{i=1}^N$ from ABC posterior 
\end{algorithmic}
\end{algorithm}

\newpage
\section{Experimental Details}\label{experimental details}
We use the adaptive SMC ABC algorithm \cite{drovandi2011estimation} in the preconditioning step for all experiments. We set the tuning parameters as $a = 0.5$, $c = 0.01$ and use 1k particles for the algorithm. As the stopping rule, we set the target MCMC acceptance rate at $10\%$, unless otherwise specified. For the unconditional density estimator, we employ unconditional normalizing flows using the \texttt{Pyro} package \cite{bingham2019pyro}, with a spline coupling layer using the transformation:
\begin{align}
    Y_{1:d} &= g_{\Tilde{\theta}}(X_{1:d}) \\
    Y_{(d+1):D} &= h_{\phi}(X_{(d+1):D}; X_{1:d})
\end{align}
where $\mathbf{X}$ are the inputs, $\mathbf{Y}$ are the outputs, e.g., $X_{1:d}$ represents the first $d$ elements of the inputs, $g_{\Tilde{\theta}}$ is either the identity function or an elementwise rational monotonic spline with parameters $\Tilde{\theta}$, and $h_{\phi}$, where $\phi$ is element-wise bijection parameter, is a conditional elementwise spline, conditioning on the first $d$ elements. Regarding the neural networks, we use four fully-connected layers and set the count bins to 16. Furthermore, if the dimensions of the parameter space are less than three, indicating a low-dimensional case, we also consider kernel density estimation with a Gaussian kernel as the unconditional density estimator, as implemented in the \texttt{Scikit-learn} package \cite{pedregosa2011scikit}. For APT and TSNPE, we use the implementation of the \texttt{sbi} package \cite{tejero2020sbi} with default settings. 

For SNPE and TSNPE, we use conditional neural spline flows \cite{durkan2019neural}. We use five coupling layers, with each coupling layer using a multilayer perceptron of two layers with 50 hidden units. The flow is trained using the Adam optimiser with a learning rate of $5 \times 10^{-4}$ and a batch size of 256. Flow training is stopped when either the validation loss, calculated on 10$\%$ of the samples, has not improved over 50 epochs or when the limit of 500 epochs is reached.

\subsection{High-dimensional SVAR model}\label{hdsvar setting}
We consider our illustrative example in a high-dimensional setting with $k = 20$, which leads to 21 parameters that need to be estimated. We set the true parameter values as follows:
\begin{align*}
    \theta &= (-0.2764,-0.7765,0.8231,-0.1972,-0.2254,0.6334,0.4495, \\
        &\qquad 0.4465,-0.8961,0.0647,-0.1791,0.0795,-0.5464,-0.9354,\\
        &\qquad -0.4639,-0.7851,-0.6833,-0.1408,0.7032,0.8321,0.1000),
\end{align*}
and use these values to simulate the observation dataset. We employ the same summary statistics as in the low-dimensional example and use a uniform  distribution as the prior, constrained between -1 and 1 for the $k$ parameters and between 0 and 1 for $\sigma$. 

Since the BSL method is the gold standard for this example, we use the standard BSL method proposed by \cite{price2018bayesian}, with the total number of simulations set to 20 million.

\subsection{BVCBM}\label{BVCBM details}
The BVCBM simulation begins by initializing a square domain with cells arranged in a hexagonal lattice. The cell at the center of the domain is identified as a cancer cell, while the others are designated as healthy cells. The simulation proceeds until the tumor reaches a volume of $100 \, {mm}^2$, in accordance with experimental measurements \citep{wade2019fabrication, kim2011active}. When this volume is attained, the distribution of healthy and cancerous cells within the lattice is recorded. This configuration then serves as the starting point to simulate tumor growth over the desired number of days. The model progresses by determining whether a cancer cell proliferates at each timestep, as described by the equation:
\begin{equation}
    p_d = p_0 \left( 1 - \frac{d}{d_{\mathrm{max}}}\right),
\end{equation}
where $p_d$ is the probability of cell division, $p_0$ is the initial division rate, $d$ is the current cell density, and $d_{\mathrm{max}}$ is the maximum density. For cancer cells that do not proliferate, the model assesses their potential to transition into invasive cells, governed by the probability $p_{\mathrm{psc}}$. Subsequently, the positions of all cells, both healthy and cancerous, are updated using Hooke's law:
\begin{equation}
    \boldsymbol{r}_i(t+\Delta t) = \boldsymbol{r}_i(t) + \frac{1}{\mu}\boldsymbol{\mathrm{F}}_i(t)\Delta t = \boldsymbol{r}_i(t) + \lambda \sum_{\forall j} \frac{\boldsymbol{r}_{i,j}(t)}{\|\boldsymbol{r}_{i,j}(t)\|}(s_{i,j}(t) - \|\boldsymbol{r}_{i,j}(t)\| ).
\end{equation}
Here, $\boldsymbol{r}_i(t+\Delta t)$ denotes the updated position of cell $i$, $\mu$ is the cell motility coefficient, $\boldsymbol{\mathrm{F}}_i(t)$ is the force on cell $i$, $\lambda$ is a mechanical interaction coefficient, $\boldsymbol{r}_{i,j}(t)$ is the vector between cells $i$ and $j$, and $s_{i,j}(t)$ is the natural length of the spring connecting the two cells. The parameters for the mechanical interactions, such as $\lambda$ and $\mu$, are sourced from prior studies in the literature \cite{meineke2001cell}. See \citet{jenner2020enhancing,wang2022calibration} for more detailed model simulation.

Four parameters $\theta = (p_0, p_{\mathrm{psc}}, d_{\mathrm{max}}, g_{\mathrm{age}})$ control the tumor growth during a single phase, which is a period when the tumor grows based on fixed values for these four parameters. For the biphasic model, an additional parameter $\tau$ is introduced, representing the time at which the tumor growth pattern changes, that is, the values for $\theta$ change. Therefore, for BVCBM, we need to estimate nine parameters for two pancreatic cancer datasets, denoted as $\theta_1 = (p_0^1,p_{\mathrm{psc}}^1,d_{\mathrm{max}}^1,g_{\mathrm{age}}^1)$, $\theta_2 = (p_0^2,p_{\mathrm{psc}}^2,d_{\mathrm{max}}^2,g_{\mathrm{age}}^2)$, and $\tau$, so that $\theta = (\theta_1, \theta_2, \tau)$.

The parameter $p_{\mathrm{psc}}$, which is the probability of tumor cell invasion into healthy cells, significantly affects the simulation time. The value of $p_{\mathrm{psc}}$ should be around $10^{-5}$, indicating that an increase in probability will require more cells to be simulated. Moreover, a smaller value of $p_{\mathrm{psc}}$ results in simulation time. For PNPE, the total simulation time for 15k simulations (26-day dataset) and 17k simulations (32-day dataset) for the preconditioning step is approximately 13 and 15 minutes, respectively, whereas SNPE requires around 1 hour for the first round (i.e.\ based on samples from the prior) of 10k simulations.  This is because the preconditioning step is effective at quickly eliminating values of $p_{\mathrm{psc}}$ that lead to longer model simulation times.

\newpage
\section{Further experimental results}\label{ferbvcbm}
\begin{figure}[ht]
\begin{center}
\centerline{\includegraphics[width=.9\columnwidth]{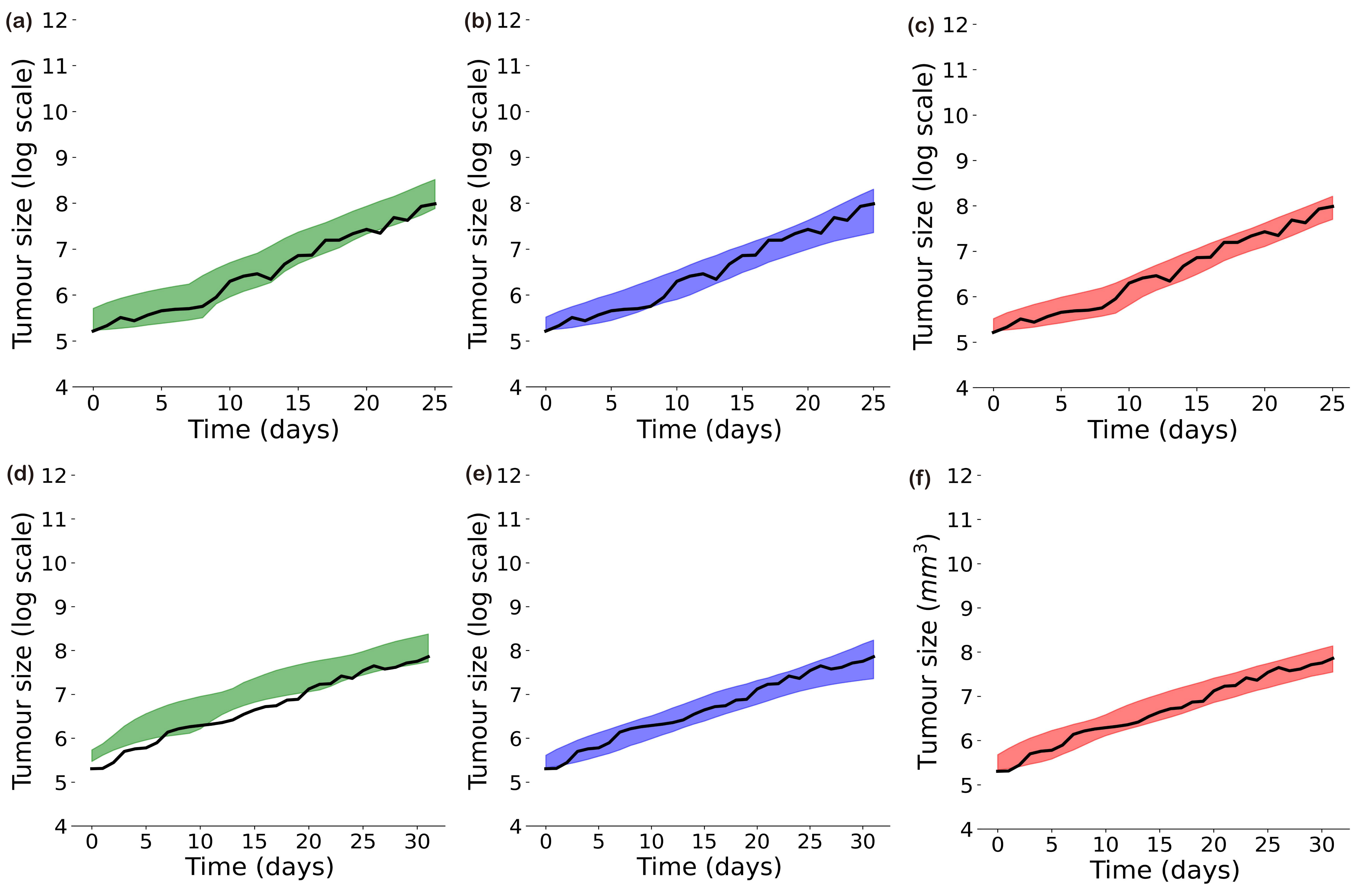}}
\caption{ \textbf{Posterior predictive distributions for two pancreatic cancer datasets in log scale.} 90\% Posterior predictive interval plots for SNPE (\textbf{a}), preconditioning step (\textbf{b}) and PNPE (\textbf{c}) for the 26-day dataset; 90\% Posterior predictive interval plots for SNPE (\textbf{d}), preconditioning step (\textbf{e}) and PNPE (\textbf{f}) for the 32-day dataset. }
\label{fig:BCVBM_pos_predictive_log}
\end{center}
\end{figure}

\begin{figure}[ht]
\begin{center}
\centerline{\includegraphics[width=.9\columnwidth]{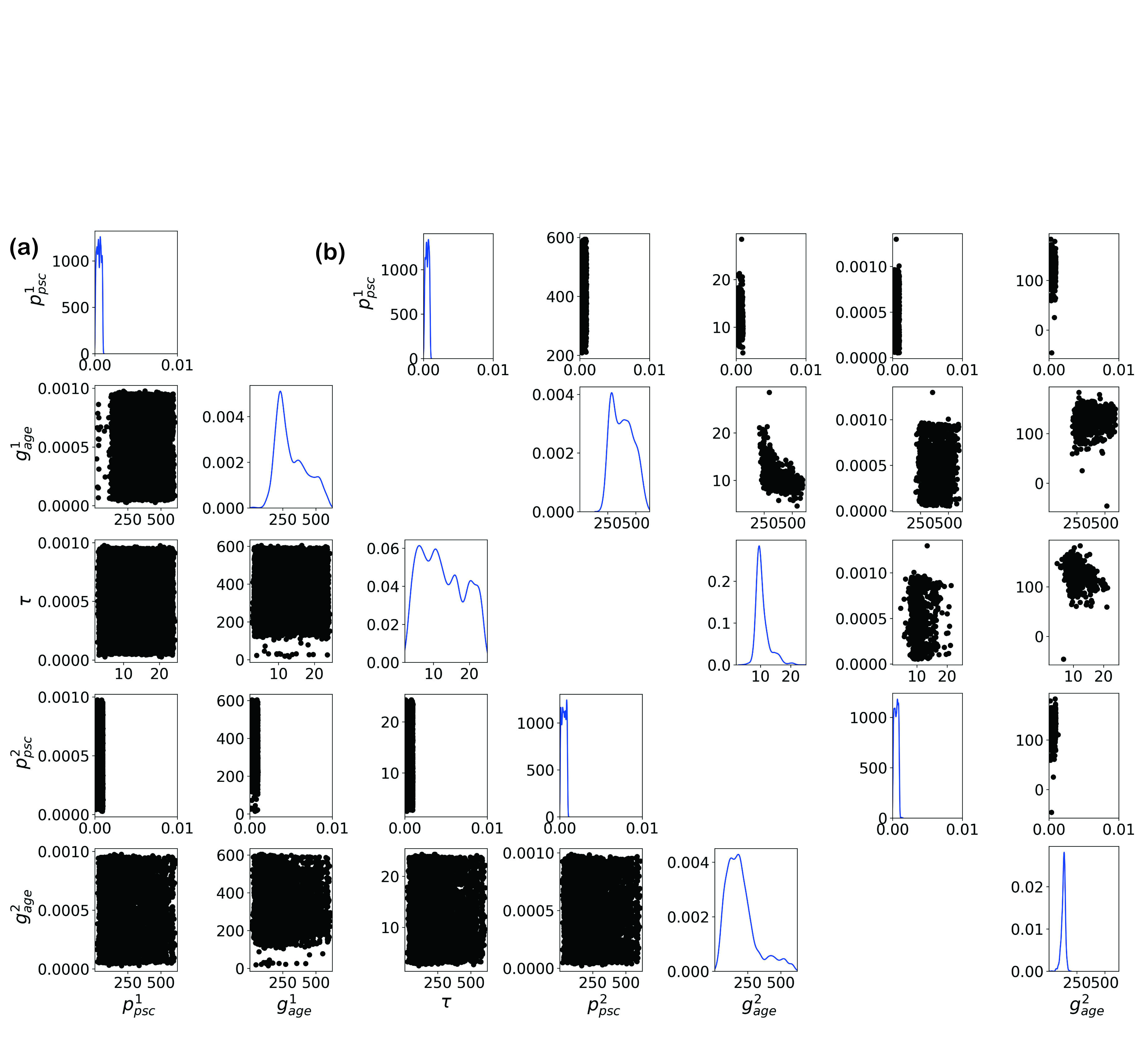}}
\caption{Bivariate density plots for the pancreatic dataset with 26-day measurements for (a) SNPE and (b) PNPE. The diagonal entries represent the marginal posterior densities for $(p_{\mathrm{psc}}^1, g_{\mathrm{age}}^1, \tau, p_{\mathrm{psc}}^2, g_{\mathrm{age}}^2)$.}
\label{fig:BCVBM_bivariate_day26}
\end{center}
\end{figure}

\begin{figure}[ht]
\begin{center}
\centerline{\includegraphics[width=.9\columnwidth]{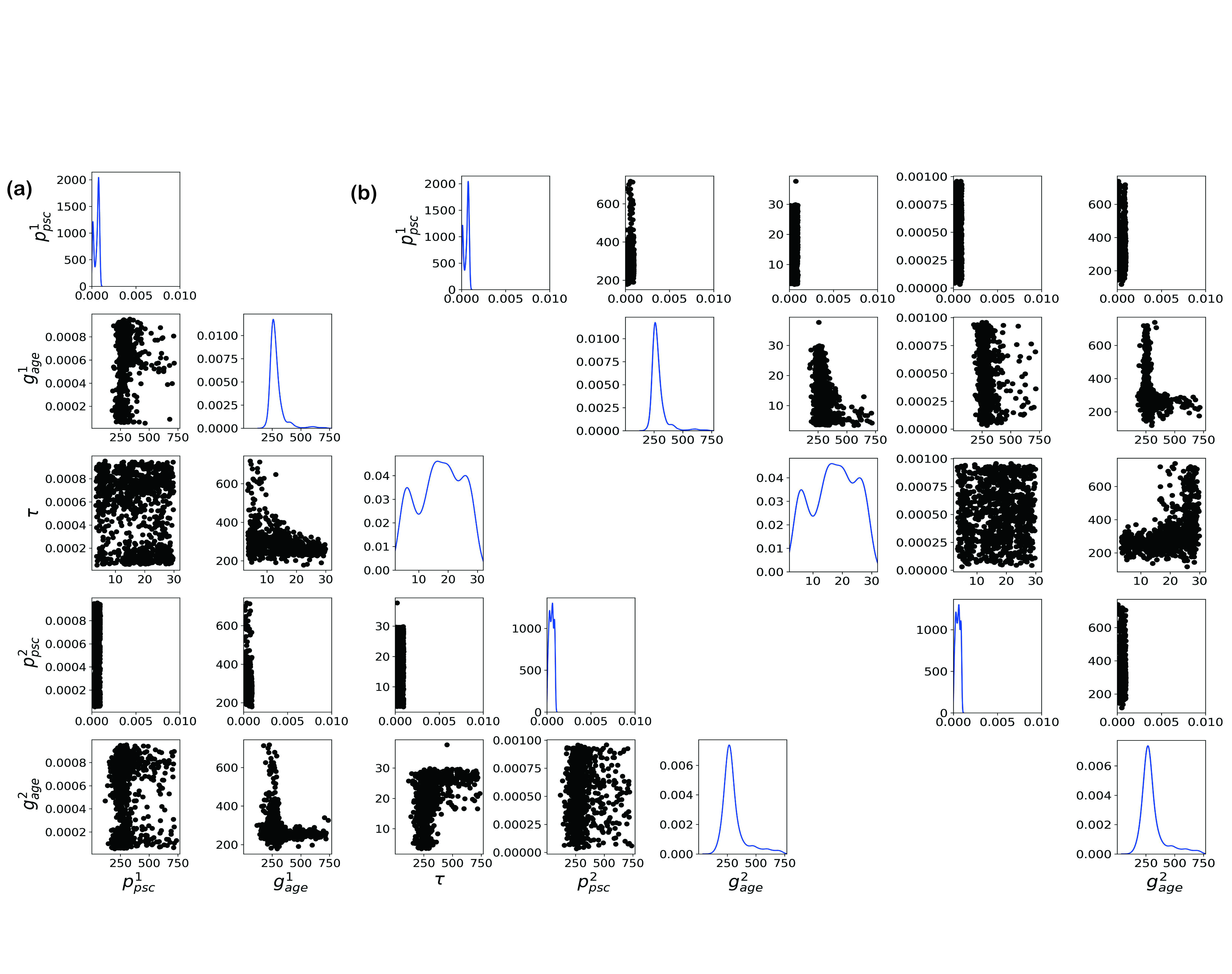}}
\caption{Bivariate density plots for the pancreatic dataset with 32-day measurements for (a) SNPE and (b) PNPE. The diagonal entries represent the marginal posterior densities for $(p_{\mathrm{psc}}^1, g_{\mathrm{age}}^1, \tau, p_{\mathrm{psc}}^2, g_{\mathrm{age}}^2)$. }
\label{fig:BCVBM_bivariate_day32}
\end{center}
\end{figure}

In this section, we present prior predictive distributions of tumour volumes in (a) and (b) for two pancreatic datasets (in (c) and (d) we show the same plots but on the log scale). 
 We also show the posterior predictive distributions on the log scale obtained with different methods in Figure \ref{fig:BCVBM_pos_predictive_log}. Then we present the bivariate posterior density and marginal posterior density plots for the BVCBM as additional results.  It is evident from Figures \ref{fig:BCVBM_bivariate_day26} and \ref{fig:BCVBM_bivariate_day32} that PNPE provides more precise estimation than SNPE for both pancreatic cancer datasets.


\begin{figure}[ht]
\begin{center}
\centerline{\includegraphics[width=.9\columnwidth]{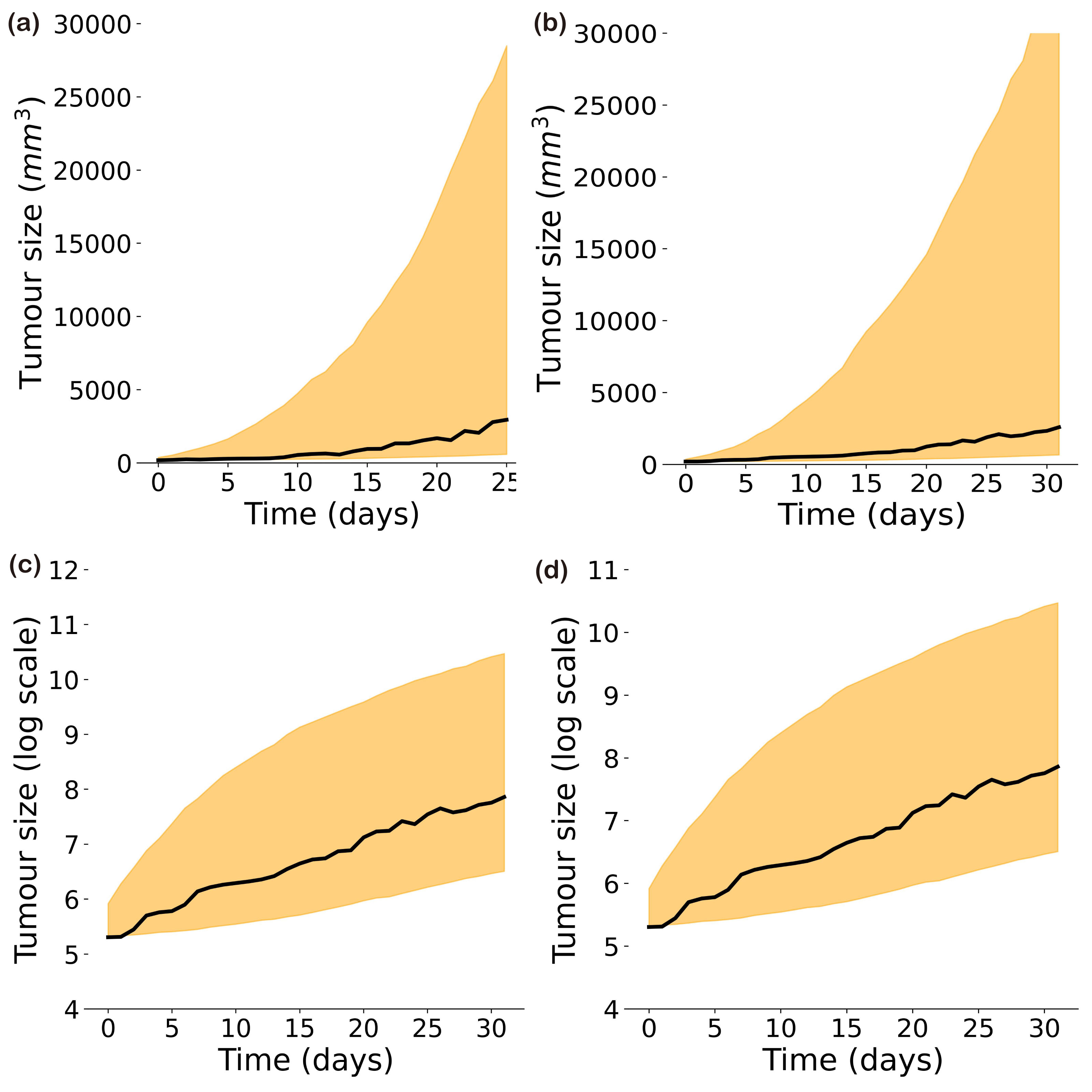}}
\caption{ \textbf{Prior predictive distribution for BVCBM.} We sample 10k parameter values from prior distribution and plot the prior predictive distribution for two pancreatic datasets. In (a) and (b), the plots are in regular scale, and in (c) and (d), the plots are in log scale. }
\label{fig:BCVBM_prior_predictive}
\end{center}
\end{figure}

\end{document}